\documentclass[journal]{IEEEtai}

\usepackage[colorlinks = true,
bookmarks = false,
linkcolor=red,
citecolor=blue,
urlcolor=blue,
anchorcolor = blue]{hyperref}

\usepackage{color,array}

\usepackage{cellspace}
\usepackage{cite}
\usepackage{amsmath,amssymb,amsfonts,nccmath}

\usepackage{algorithmic}
\usepackage{graphicx,balance}
\usepackage{subfig}
\usepackage{balance}
\usepackage{textcomp}
\usepackage{xcolor}
\usepackage{color}
\usepackage{url}
\usepackage{xspace}
\usepackage{booktabs}
\usepackage{comment}
\usepackage{acronym}
\usepackage[none]{hyphenat}
\usepackage{ragged2e}
\usepackage{tabularx}
\usepackage{tablefootnote}

\usepackage{colortbl} 
 \usepackage[normalem]{ulem}
 \useunder{\uline}{\ul}{}

\newcommand{\eg}{e.g.,\xspace}
\newcommand{\ie}{i.e.,\xspace}
\newcommand{\etal}{et\ al.\xspace}
\newcommand{\etc}{etc.\@\xspace}

\acrodef{AI}{Artificial Intelligence}
\acrodef{XAI}{Explainable AI}
\acrodef{ML}{Machine Learning}
\acrodef{DL}{Deep Learning}
\acrodef{NLP}{Natural Language Processing}
\acrodef{LLMs}{Large Language Models}
\acrodef{GPT}{Generative Pre-trained Transformer}
\acrodef{BERT}{Bidirectional Encoder Representations from Transformers}
\acrodef{GANs}{Generative Adversarial Networks}
\acrodef{VAEs}{Variational Autoencoders}
\acrodef{GPUs}{Graphics Processing Units}
\acrodef{RNNs}{Recurrent Neural Networks}
\acrodef{CNNs}{Convolutional Neural Networks}
\acrodef{ReLU}{Rectified Linear Unit}
\acrodef{FFN}{Position-Wise Feed Forward Network}
\acrodef{ELBO}{Evidence Lower Bound}
\acrodef{KL}{Kullback-Leibler}
\acrodef{DCGANs}{Deep Convolutional GANs}
\acrodef{WGANs}{Wasserstein GANs}
\acrodef{NLU}{Natural Language Understanding}
\acrodef{GPUs}{Graphics Processing Units}
\acrodef{TPUs}{Tensor Processing Units}
\acrodef{LSTM}{Long Short-term Memory}
\acrodef{NLG}{Natural Language Generation}
\acrodef{GELU}{Gaussian Error Linear Unit}
\acrodef{RL}{Reinforcement Learning}
\acrodef{RLHF}{Reinforcement Learning from Human Feedback}
\acrodef{DPO}{Direct Policy Optimization}
\acrodef{NCE}{Noise-Contrastive Estimation}
\acrodef{DSM}{Denoising Score Matching}
\acrodef{MoE}{Mixture of Experts}
\acrodef{HMMs}{Hidden Markov Models}
\acrodef{DNNs}{Deep Neural Networks}
\acrodef{ASR}{Automatic Speech Recognition}
\acrodef{MLM}{Masked Language Modeling}
\acrodef{CLIP} {Contrastive Language-Image Pre-training}
\acrodef{ViT}{Vision Transformer}

\renewcommand\IEEEkeywordsname{Keywords}
\begin{document}

\title{Recent Advances in Generative AI and Large Language Models: Current Status, Challenges, and Perspectives} 

%\begin{comment}
\author{Desta Haileselassie Hagos, \IEEEmembership{Member, IEEE}, Rick Battle, and Danda B. Rawat, \IEEEmembership{Senior Member, IEEE} \vspace{-0.99ex}
\thanks{This work was supported by the United States DoD Center of Excellence in AI/ML at Howard University under Contract number W911NF-20-2-0277 with the U.S. Army Research Laboratory (ARL). However, any opinions, findings, conclusions, or recommendations expressed in this document are those of the authors and should not be interpreted as representing the official policies, either expressed or implied, of the funding agencies.}
\thanks{D. H. Hagos and D. B. Rawat are with the DoD Center of Excellence in Artificial Intelligence and Machine Learning (CoE-AIML), College of Engineering and Architecture (CEA), Department of Electrical Engineering and Computer Science, Howard University, Washington DC, USA (e-mail: desta.hagos@howard.edu; danda.rawat@howard.edu).}
\thanks{R. Battle is with the VMware AI Labs by Broadcom, Palo Alto, CA, USA. (e-mail: rick.battle@broadcom.com).} 
%\thanks{This paragraph will include the Associate Editor who handled your paper.}
}
%\end{comment}

\markboth{Accepted at IEEE Transactions on Artificial Intelligence (TAI)}
{Hagos \MakeLowercase{\textit{et al.}}: Recent Advances in Generative AI and Large Language Models: Current Status, Challenges, and Perspectives}

\maketitle

\begin{abstract}
The emergence of Generative \ac{AI} and \ac{LLMs} has marked a new era of \ac{NLP}, introducing unprecedented capabilities that are revolutionizing various domains. This paper explores the current state of these cutting-edge technologies, demonstrating their remarkable advancements and wide-ranging applications. Our paper contributes to providing a holistic perspective on the technical foundations, practical applications, and emerging challenges within the evolving landscape of Generative \ac{AI} and \ac{LLMs}. We believe that understanding the generative capabilities of \ac{AI} systems and the specific context of \ac{LLMs} is crucial for researchers, practitioners, and policymakers to collaboratively shape the responsible and ethical integration of these technologies into various domains. Furthermore, we identify and address main research gaps, providing valuable insights to guide future research endeavors within the \ac{AI} research community. 
\end{abstract}

\begin{IEEEImpStatement}
Understanding the full potentials and limitations of Generative \ac{AI} and \ac{LLMs} shapes the future of \ac{NLP} and its impact on various industries and societies. This paper explores the transformative potential of advanced \ac{NLP} tools like Generative \ac{AI} and \ac{LLMs}, shaping the future of communication and understanding across diverse domains. Our paper not only addresses the current state of Generative \ac{AI} and \ac{LLMs} in language understanding, machine translation, question answering, text summarization, and code Completion but also makes a significant contribution in addressing some of the critical research gaps of Generative \ac{AI} and \ac{LLMs}. By addressing issues of bias and fairness, interpretability, fine-tuning and adaptability, domain adaptation, data privacy and security, computational cost, deepfake generation, human-AI collaboration, long-term planning, limited context window, and long-term memory, \etc, our work aims to pave the way for responsible, ethical, and impactful integration of these transformative technologies across diverse domains. We believe that this research serves as a roadmap for the AI community, pushing towards an ethical, inclusive, and impactful future. It empowers diverse domains with transformative technologies, creating a robust landscape for the responsible evolution of \ac{AI}.

\end{IEEEImpStatement}


%\begin{comment}
\begin{table}[!t]
\centering
\caption*{List of acronyms used in this paper.}
\def\arraystretch{1.13}
\begin{tabular}{l|l}
\toprule
\textbf{Acronym} & \hspace*{\fill} \textbf{Definition} \hspace*{\fill} \\ \midrule %\hline
%\hline
AI & Artificial Intelligence \\ %\hline
ASR & Automatic Speech Recognition \\
BERT & Bidirectional Encoder Representations from Transformers \\
CLIP & Contrastive Language-Image Pre-training \\
CNNs & Convolutional Neural Networks \\
DCGANs & Deep Convolutional GANs \\
DL & Deep Learning \\
DNNs & Deep Neural Networks \\
DPO & Direct Policy Optimization \\
DSM & Denoising Score Matching \\
ELBO & Evidence Lower Bound \\
FFN & Position-Wise Feed Forward Network \\
GANs & Generative Adversarial Networks \\ 
GELU & Gaussian Error Linear Unit \\
GPT & Generative Pre-trained Transformer \\
GPUs & Graphics Processing Units \\
HMMs & Hidden Markov Models \\
KL & Kullback-Leibler \\
LLMs & Large Language Models \\
LSTM & Long Short-term Memory \\
ML & Machine Learning \\
MLM & Masked Language Modeling \\
MoE & Mixture of Experts \\
NCE & Noise-Contrastive Estimation \\
NLG & Natural Language Generation \\
NLP & Natural Language Processing \\
NLU & Natural Language Understanding \\
ReLU & Rectified Linear Unit \\
RL & Reinforcement Learning \\
RLHF   & Reinforcement Learning from Human Feedback \\
RNNs & Recurrent Neural Networks \\
TPUs & Tensor Processing Units \\
XAI & Explainable Artificial Intelligence \\ 
VAEs & Variational Autoencoders \\ 
ViT & Vision Transformer\\
WGANs & Wasserstein GANs \\

\bottomrule

\end{tabular}

\end{table}
%\end{comment}

\section{Introduction}
\label{introduction}

\IEEEPARstart{I}{n} today's data-driven world, the ability to effectively process and understand natural language is becoming increasingly important. Generative \ac{AI} and \ac{LLMs} have emerged as powerful tools that are expanding the boundaries of \ac{NLP}, offering unprecedented capabilities across a variety of domains. \ac{LLMs}, being a specific application of Generative \ac{AI}, play a foundational role in the broader landscape of generative capabilities of \ac{AI}, demonstrating remarkable abilities in understanding and generating human language, opening up many opportunities across a wide range of domains. Their ability to process and analyze vast amounts of text data has enabled them to tackle complex linguistic tasks such as machine translation~\cite{kalchbrenner2013recurrent, papineni2002bleu}, text summarization~\cite{lewis2019bart}, question answering~\cite{brown2020language}, mathematical reasoning~\cite{cobbe2021training}, and code generation~\cite{chen2021evaluating} with unprecedented accuracy~\cite{brants2007large}. Recent \ac{AI} advancements have revolutionized our ability to understand, create, and engage with human language~\cite{manning2022human, brown2020language}. Overcoming the challenges related to understanding and generating human language has been one of the main goals of \ac{AI} research. This progress has been made possible through the development of new state-of-the-art \ac{LLMs} and Generative \ac{AI} models. This rapid advancement is the result of several factors, some of which are listed below.

\vspace{1.0ex}
\noindent \textbf{Advances in Computational Power}. The explosion of data and the increasing computational power accessible to researchers, organizations, and companies has enabled the training of complex neural networks~\cite{svyatkovskiy2017training}. As computational power has increased, larger and more complex neural networks have become possible, leading to the development of \ac{LLMs} and Generative \ac{AI} models that can perform tasks that were previously impossible, such as generating realistic text and images. These powerful computing resources are essential for processing and modeling the vast amount of data required to train LLMs and generative AI, enabling them to learn the patterns and relationships necessary for their tasks. The development of powerful new computing hardware, such as \ac{GPUs}, has facilitated the training of \ac{AI} models on massive datasets of text and code~\cite{li2014large}. The increasing availability of computational power has also reduced the time and cost of training \ac{LLMs} and generative \ac{AI} models, making it more feasible for researchers and companies to develop and deploy them~\cite{isaev2023scaling}.

\vspace{1.0ex}
\noindent \textbf{Datasets Availability and Scale}. The increasing availability of data has enabled the training of \ac{LLMs} and Generative \ac{AI} models on larger and more diverse datasets, significantly improving their performance~\cite{kaplan2020scaling}. The vast amounts of text, audio, images, and video content produced in the digital age provide valuable resources for training AI models, which rely on these massive datasets to learn the complexities of human language and content creation. The work in~\cite{kaplan2020scaling} indicates that dataset size is a key factor in determining the performance of \ac{LLMs} and that larger datasets lead to significant improvements in model performance. In~\cite{hoffmann2022empirical} a more efficient approach to training LLMs is proposed in terms of computation and data usage. The authors suggest that for optimal LLM scaling, it is essential to equally scale the model size and training dataset size. This implies that having a sufficiently large dataset is vital for achieving the best performance.

\vspace{1.0ex}
\noindent \textbf{Deep Learning Advances}. New \ac{ML} algorithms, such as \ac{DL}, have been developed that can learn complex patterns from data. Deep learning techniques, especially deep neural networks with many layers, have made remarkable advancements~\cite{lecun2015deep}. Innovations like \ac{RNNs}~\cite{medsker2001recurrent, socher2013recursive}, \ac{CNNs}~\cite{lecun1995convolutional}, and Transformers~\cite{vaswani2017attention} have paved the way for more advanced and capable models. The Transformer architecture, in particular, played a significant role in the development of \ac{LLMs}~\cite{vaswani2017attention}.

\vspace{1.0ex}
\noindent \textbf{Transfer Learning and Pre-training}. \ac{LLMs} are trained on massive datasets of text, giving them a broad understanding of the world and how language is used. For example, the \ac{GPT}-3 language model was trained on a dataset of 175 billion words~\cite{brown2020language}. Transfer learning plays a critical role in the development of highly efficient and effective \ac{LLMs} and generative AI models~\cite{raffel2020exploring}. Models like \ac{BERT}~\cite{devlin2019bert}, \ac{GPT}~\cite{radford2018improving}, and their variants are pre-trained on massive text corpora, giving them a broad understanding of language. This pre-trained knowledge can be leveraged for various downstream tasks without the need for retraining the model from scratch, which can be both computationally expensive and time-consuming~\cite{raffel2020exploring}. Transfer learning enables the use of pre-trained models that have already been trained on a large dataset. This reduces the amount of training data that we need for our specific task. For example, if we want to train a model to translate text from English to Chinese, we can fine-tune a pre-trained language model that was trained on a dataset of English and Chinese text. This approach is particularly useful in scenarios where obtaining large labeled datasets is challenging and expensive since it reduces the amount of training data that we need to collect and label. Transfer learning significantly reduces the computational and data requirements for developing effective language models.  Instead of training a separate model for each specific task, a pre-trained language model can be fine-tuned on a smaller task-specific dataset. This fine-tuning process is faster and requires less data, making it a practical approach for a wide range of applications~\cite{houlsby2019parameter}.

\vspace{1.0ex}
\noindent \textbf{Modern Neural Network Architectures}. The emergence of neural network architectures, such as the \ac{GPT}~\cite{radford2018improving} and \ac{VAEs}~\cite{kingma2013auto}, has led to the development of modern \ac{LLMs} and generative AI. \ac{LLMs} need to be able to learn long-range dependencies in text to generate coherent and meaningful text in a variety of formats~\cite{brown2020language}. Traditional \ac{RNNs}~\cite{socher2013recursive}, \eg \ac{LSTM}, are not well-suited for this task because they have difficulty learning long-range dependencies beyond a few words. However, the \emph{transformer} architecture can learn long-range dependencies more effectively~\cite{radford2018improving}. The work in~\cite{vaswani2017attention} demonstrates that the \emph{transformer} architecture outperformed \ac{RNNs} on a variety of \ac{NLP} tasks, including machine translation and text summarization~\cite{papineni2002bleu, feng2018attentive}.

\vspace{1.0ex}
\noindent \textbf{Community Collaboration and Open-Source Initiatives}. The \ac{AI} research community, through collaborative efforts and open-source initiatives, such as OpenAI~\cite{brown2020language}, Hugging Face~\cite{wolf2020transformers}, Google AI~\cite{thoppilan2022lamda}, \etc, has significantly contributed to the advancement of state-of-the-art \ac{LLMs} and Generative \ac{AI}. This progress is the result of joint collaboration among AI researchers and developers from various organizations and research institutions. These collaborations have facilitated the sharing of knowledge, expertise, and resources, enabling rapid progress. The open-source movement has played a critical role in accelerating the development of \ac{LLMs} and Generative \ac{AI}. By making source codes, data, and models publicly available, open-source initiatives have allowed researchers and developers to build upon each other's work, leading to faster innovation and more robust models. Open-source platforms like Hugging Face and GitHub serve as hubs for sharing pre-trained models, datasets, and fine-tuning scripts. Additionally, open-source projects and community efforts have made substantial corpora of text data available for training robust language models, such as Wikipedia, Common Crawl, and Project Gutenberg.

\vspace{1.5ex}
\noindent \textbf{Contributions}. In this paper, we make the following main contributions. 

\vspace{1.0ex}
\begin{itemize}
    \item We provide a holistic perspective on the current landscape of the generative capabilities of \ac{AI} systems and the specific context of \ac{LLMs}.
    \item We demonstrate the significant progress and unprecedented capabilities introduced by the emergence of Generative \ac{AI} and \ac{LLMs}.
    
    \item We provide valuable insights that can guide future research endeavors within the AI research community.
    \item Finally, we identify and address several key research gaps in the field of Generative \ac{AI} and LLMs.

\end{itemize}

\vspace{1.5ex}
\noindent \textbf{Organization}. The rest of this paper is organized as follows. Section~\ref{generative_AI} introduces the overview of generative models and explores the applications of Generative \ac{AI}. In Section~\ref{language_modeling}, we discuss the traditional and modern approaches to language modeling and the applications of \ac{LLMs}. Section~\ref{challenges_of_generative_AI_and_LLMs} provides a detailed discussion of the challenges associated with Generative \ac{AI} and \ac{LLMs} and potential solutions. The impact of identified research gaps and future directions on the ethical and responsible integration of Generative \ac{AI} and \ac{LLMs} is presented in Section~\ref{future_directions}. Finally, Section~\ref{conclusion} concludes our paper and suggests directions for future research work.

\section{Generative AI}
\label{generative_AI}

Generative \ac{AI} refers to a class of algorithms and models within \ac{AI} and \ac{NLP} that are designed to generate new, previously unseen data that is similar to existing examples by employing a variety of techniques~\cite{radford2018improving}. These models learn the underlying patterns and structures present in the training data and use that knowledge to create novel instances that resemble the original data. It has the potential to revolutionize many industries and creative fields. Generative \ac{AI} models are trained on large datasets of existing content. Generative models aim to capture the underlying distribution of data, enabling them to generate new samples that are statistically similar to the training data. To achieve this, generative models employ a latent space, denoted as $Z$, which represents a hidden or underlying representation of the data. This latent space is then mapped to the actual data space, denoted as $X$, through a generator function, represented by $G_\theta(Z)$. The parameter $\theta$ represents the adjustable parameters of the generative model, which are optimized during the training process. The goal of training a generative model is to make the generated samples, $G_\theta(Z)$, virtually indistinguishable from real data samples by focusing on maximizing the probability of generating the observed data samples. The objective function for training a generative model, without specifying a particular architecture, is expressed in Equation~\ref{eq:generative_model}, where $N$ is the number of training samples, $x^{(i)}$ represents the $i^{th}$ training sample, and $p_{\text {model }}\left(x^{(i)} ; \theta\right)$ denotes the probability assigned by the generative model to the $i^{th}$ training sample.

\begin{equation}
\max _\theta \sum_{i=1}^N \log p_{\text {model }}\left(x^{(i)} ; \theta\right)
\label{eq:generative_model}
\end{equation}

\subsection{\ac{GANs}}

\ac{GANs} are a type of generative AI model that consists of two neural networks: a generator and a discriminator~\cite{creswell2018generative}. The generator is responsible for creating new realistic and high-quality data, including images, text, and music, by learning the underlying distribution of the data~\cite{goodfellow2014generative}. The discriminator, on the other hand, is responsible for distinguishing whether the new data is real or fake~\cite{goodfellow2014generative}. The fundamental principle behind \ac{GANs} involves a generator network creating realistic data, such as images, and a discriminator network evaluating the generated data by distinguishing between real and fake data~\cite{goodfellow2014generative}. Over time, the generator improves its ability to create realistic data by attempting to deceive the discriminator, which enhances its ability to distinguish between real and generated data~\cite{goodfellow2014generative}. The training process of a GAN network, as shown in Equation~\ref{eq:GANs}, involves optimizing the parameters of both the generator (represented by $G$) and discriminator (represented by $D$) networks~\cite{goodfellow2014generative}. Here, $p_{{data }}(x)$ denotes the distribution of real data, $p_z(z)$ represents the distribution of random noise in the latent space, $x$ denotes a real data point, $G(z)$ is a data point generated from random noise $z$, $D(x)$ is the discriminator's output indicating the probability that $x$ is real, and \textit{log} refers to the natural logarithm. The objective is to minimize the log-probability of the discriminator correctly identifying whether a sample is real or generated, while simultaneously maximizing the log-probability of the generator producing data that the discriminator perceives as real.

\begin{multline}
\min_G \max_D V(D, G)=\\
\mathbb{E}_{x \sim p_{\text {data }}(x)}[\log D(x)]+\mathbb{E}_{z \sim p_z(z)}[\log (1-D(G(z)))]
\label{eq:GANs}
\end{multline}

Within the adversarial setting, various classes of \ac{GANs} have emerged over the years, each tailored to specific tasks in the generative modeling space. For example, the work of Radford \etal~\cite{radford2015unsupervised} presents a \ac{DCGANs} by extending the \ac{GANs} architecture, an extension of the original GAN architecture proposed by Goodfellow \etal~\cite{goodfellow2014generative}. \ac{DCGANs} employ \ac{CNNs} in both the generator and discriminator, enabling the generation of high-quality images. \ac{CNNs} are known to perform well at capturing spatial relationships in data~\cite{radford2015unsupervised}, making them well-suited for image generation tasks. Addressing the training instability issues of~\cite{goodfellow2014generative}, Arjovsky \etal introduced the \ac{WGANs} algorithm~\cite{arjovsky2017wasserstein}. \ac{WGANs} replace the binary cross-entropy loss with the Wasserstein distance, leading to improved stability and convergence during training~\cite{arjovsky2017wasserstein}. In the context of \ac{GANs}, the Wasserstein distance defines the objective function between two distributions, denoted as $A$ and $B$, as shown in Equation~\ref{eq:WGANs}. Here, $W(A, B)$ represents the Wasserstein distance between distributions $A$ and $B$, $\inf$ denotes the infimum, which represents the minimum value, $\gamma$ refers to a joint distribution defined on the product space of $A$ and $B$, $\Pi(A, B)$ is the set of all joint distributions with marginals $A$ and $B$. The terms $(x, y$ represent samples from the joint distribution $\gamma$, and $d(x, y)$ denotes the distance between $x$ and $y$ in the metric space.

\begin{equation}
W(A, B)=\inf _{\gamma \in \Pi(A, B)} \mathbb{E}_{(x, y) \sim \gamma}[d(x, y)]
\label{eq:WGANs}
\end{equation}

To tackle the challenges associated with training high-resolution \ac{GANs}, Karras \etal~\cite{karras2017progressive} proposed a progressive growth of \ac{GANs}. This algorithm employs a progressive training strategy that gradually increases the resolution of the generated images throughout the training process. This approach allows the algorithm to capture finer details and produce high-resolution images with enhanced stability and scalability~\cite{karras2017progressive}.

\subsection{Variational Autoencoder Models}

\ac{VAEs} are generative models that learn a probabilistic mapping from the data space to a latent space, a lower-dimensional representation of the data that captures its essential features, enabling the generation of new samples through sampling from the learned latent space~\cite{kingma2013auto}. This process involves two key components: encoders and decoders. In the \ac{VAEs} framework, encoders and decoders, play important roles in the process of learning and generating data. The encoder is implemented using a neural network and it is responsible for mapping the input data $x$ to a probability distribution in the latent space  $z$, as shown in Equation~\ref{eq:encoder}. Similar to the encoder, the decoder is also implemented using a neural network, and it reconstructs the original data from this latent representation, $z$, as illustrated in Equation~\ref{eq:decoder}.  The encoder and decoder are trained jointly using a technique called variational inference~\cite{kingma2013auto, kingma2016improved}. Variational inference minimizes two losses: a reconstruction loss and a regularization loss.  In Equation~\ref{eq:encoder} $\mu_\phi(x)$ and $\sigma_\phi(x)$ represent the mean and standard deviation of the distribution, respectively. In Equation~\ref{eq:decoder}, the parameters $\mu_\theta(z)$ and $\sigma_\theta(z)$ represent the mean and standard deviation of the latent space distribution, which are learned by the decoder neural network during training.

\begin{equation}
z= q_\phi(z \mid x)=\mathcal{N}\left(\mu_\phi(x), \sigma_\phi(x)^2\right)
\label{eq:encoder}
\end{equation}

\begin{equation}
p_\theta(x \mid z)=\mathcal{N}\left(\mu_\theta(z), \sigma_\theta(z)^2\right)
\label{eq:decoder}
\end{equation}

The reparameterization trick, introduced in \ac{VAEs} to facilitate backpropagation through the sampling process~\cite{kingma2013auto}, addresses the challenge of applying backpropagation to inherently random sampling operations. While backpropagation is a fundamental algorithm for training neural networks, its direct application to sampling is problematic due to the randomness involved. The reparameterization trick provides an alternative approach to sample from a distribution while maintaining the necessary connections for backpropagation~\cite{kingma2013auto}. In \ac{VAEs}, this technique is employed to sample the latent variable, $z$, from a simple distribution, typically a standard normal distribution. These samples are then transformed to match the distribution produced by the encoder, as described in Equation~\ref{eq:repar_trick}. This transformation ensures that the sampled latent variables remain consistent with the encoder's understanding of the data while preserving the randomness required for generating new samples. In Equation~\ref{eq:repar_trick}, the $\epsilon$ represents a random noise vector sampled from a standard normal distribution, $\odot$ represents the element-wise product operation, $\sigma_\theta(x)$ represents the standard deviation of the distribution produced by the encoder, and $\mu_\theta(x)$ represents the mean of the distribution produced by the encoder.

\begin{equation}
    z=\mu_\theta(x) + \sigma_\theta(x)\odot \epsilon, \text{~where}~ \epsilon \sim \mathcal{N}(0, 1)
    \label{eq:repar_trick}
\end{equation}

 The main objective for training a VAE is to maximize the~\ac{ELBO}~\cite{kingma2013auto, rezende2015variational}. Maximizing the \ac{ELBO} during training encourages the VAE to learn a meaningful and smooth latent space representation for the input data~\cite{kingma2013auto, rezende2015variational}. By maximizing the \ac{ELBO}, the VAE is trained to learn a latent space that captures the underlying structure of the data while also allowing for the efficient generation of new samples~\cite{kingma2013auto, rezende2015variational}. The \ac{ELBO}, as shown in Equation~\ref{eq:ELBO}, comprises two terms: the reconstruction loss of the data given the latent variable ($log p_\theta(x \mid z)$), which measures the expected log-likelihood of the data given the latent variable, and the \ac{KL} divergence between the approximate posterior (encoder) and the prior distribution ($D_{\mathrm{KL}}\left(q_\phi(z \mid x) \| p(z)\right)$). The \ac{KL} divergence encourages the latent distribution learned by the encoder to be similar to the prior distribution, which is typically a standard normal distribution. This constraint helps prevent the encoder from learning overly complex or entangled latent representations. In Equation~\ref{eq:ELBO}, the $\mathcal{L}$ denotes the overall loss function.

\begin{equation}
\mathcal{L}(\theta, \phi ; x)=\mathbb{E}_{q_\phi(z \mid x)}\left[\log p_\theta(x \mid z)\right]-D_{\mathrm{KL}}\left(q_\phi(z \mid x) \| p(z)\right)
\label{eq:ELBO}
\end{equation}

\subsection{Autoregressive Models}

In the context of Generative \ac{AI},  autoregressive models are a class of likelihood models that generate new sequential data by predicting the next value in a sequence based on the previous values. These models involve modeling the probability distribution of each element in a sequence given the entire history of previous elements, $P(x_t|x_{t1}, x_{t2},\ldots, x_1)$. This ability makes autoregressive models well-suited for a variety of \ac{NLP} tasks where the ability to understand and generate coherent sequences is essential~\cite{meek2002autoregressive}. They are also widely used in capturing the dynamics of time series data~\cite{meek2002autoregressive}. An autoregressive model of order $p$ can be generally represented as shown in Equation~\ref{eq:AR_models} where $X_t$ denotes the value of the time series at time $t$, $c$ is a constant term, $\phi_i$ are the autoregressive coefficients, representing the influence of the previous $i^{th}$ observations on the current observation, and $\epsilon_t$  is an error term, which represents the random noise in the data. The parameters of the model ($c, \phi_i$) are typically estimated from the observed data using methods like maximum likelihood estimation.

\begin{equation}
X_t = c + \sum_{i=1}^{p} \phi_i X_{t-i}+\epsilon_t
    \label{eq:AR_models}
\end{equation}

\subsection{Mixture of Expert Models}

A \ac{MoE} model represents a neural network architecture that combines the strengths of specialized expert networks with a gating mechanism to perform complex tasks~\cite{shazeer2017outrageously, wang2020deep}.  In the context of \ac{NLP} architectures, \ac{MoE} models are applied to enhance the capabilities and efficiency of the underlying language generation architecture~\cite{shazeer2017outrageously, du2022glam}. Within the realm of \ac{MoE} models, these architectures optimize resource utilization by selectively activating relevant experts for specific tasks, demonstrating adaptability to different domains through the integration of domain-specific expert models~\cite{gururangan2021demix}. Moreover, \ac{MoE} architectures offer scalability, allowing the addition of more expert networks to handle a broader range of tasks~\cite{rajbhandari2022deepspeed}. The advantages of \ac{MoE} models extend beyond their architectural complexities. Recent studies, such as the work presented in~\cite{rajbhandari2022deepspeed}, emphasize their scalability, enabling the addition of more expert networks to handle a broader range of tasks. Furthermore, these models have demonstrated the ability to achieve superior model quality compared to their dense counterparts, even with significantly reduced training costs. However, despite these advantages, \ac{MoE} models pose some critical challenges. \ac{MoE} models are sensitive to small changes in the gating network weights. Since the gating network determines the contribution of each expert to the final prediction, even slight changes in these weights can lead to significant shifts in the model's training stability and cause unpredictable behavior~\cite{shazeer2017outrageously}. This sensitivity can make training and optimization of the model more challenging. To mitigate this, techniques such as sparse routing have been proposed~\cite{zhou2022mixture, chi2022representation}. Regularization techniques such as weight decay or dropout also help mitigate sensitivity to small changes in gating network weights by preventing overfitting and promoting smoother decision boundaries~\cite{wang2020deep}. Additionally, training \ac{MoE} models can be computationally intensive, especially when dealing with a large number of experts or complex gating functions. Each forward pass through the network involves evaluating the outputs of multiple experts and updating the parameters of both the expert and gating networks. This computational overhead can make training slower and require more resources compared to simpler neural network architectures. Developing more efficient training algorithms specifically tailored for \ac{MoE} models can help reduce computational intensity. The overall \ac{MoE} model architecture can be broken down into several key components including the following.

\vspace{1.0ex}
\noindent{\textbf{Expert Networks}}. One of the main features of the \ac{MoE} model is the presence of multiple expert networks. These expert networks play a critical role in learning specific patterns or features within the input data and serve as the core models of the \ac{MoE} system. Each expert network is tailored to specialize in a particular aspect or subset of the input problem space. %For example, in an \ac{NLP} task, different expert networks might focus on diverse linguistic structures or semantic aspects~\cite{du2022glam}.

\vspace{1.0ex}
\noindent\textbf{Gating Network}. The gating network mechanism a crucial component that analyzes the input data and decides which expert network is most suitable for a given instance~\cite{zhou2022mixture}. It assigns weights to each expert, indicating their relevance or contribution to the current input. The gating network typically outputs a probability distribution over available experts, reflecting the relevance of each expert to the current input~\cite{zhou2022mixture}. There are two main types of \ac{MoE} routing strategies in \ac{MoE} systems: dense routing and sparse routing. In dense routing, every input is directed to all experts, and the final output is a weighted combination of all expert predictions based on the gating network's output. On the other hand, sparse routing is a more efficient approach where the gating network selects only a subset of experts for each input, reducing computational cost~\cite{shazeer2017outrageously, chen2022towards}. The \ac{MoE} model dynamically combines the predictions of multiple experts based on learned gating coefficients, allowing it to adaptively switch between different experts depending on the input data. This mechanism enables the model to capture complex patterns and improve performance compared to a single expert model. The gating network is generally represented as shown in Equation~\ref{gating_network:eq} where $g_k(x)$ denotes the gating function for gate $k$, $\sigma$ is an activation function (usually sigmoid or softmax), and $W_g$ represents the parameters of the gating network.

%\vspace{-1.0ex}
\begin{equation}
g_k(x)=\sigma\left(W_{g k}^T x\right)
\label{gating_network:eq}
\end{equation}

%\vspace{-1.0ex}

\vspace{1.0ex}
\noindent \textbf{Output Computation}.  When the experts are activated,  they process input data and generate individual predictions. These predictions are then combined to form the final output of the \ac{MoE} model. The specific method of combining predictions depends on the task and \ac{MoE} architecture. In the weighted averaging approach, predictions from each expert are weighted based on the output of the gating network, and the weighted average is taken as the final output. In classification tasks, experts can vote for the most likely class, and the majority vote becomes the final prediction~\cite{yuksel2012twenty}. The output of a \ac{MoE} model, denoted as $y(x)$, is computed using Equation~\ref{eq:MoE_output}, representing a weighted sum of the expert outputs. The final output, $y(x)$, is computed by aggregating the contributions of all experts. It sums up the weighted outputs of all experts based on the gating values, resulting in the \ac{MoE}'s prediction. This output is often passed through additional layers, such as fully connected layers or activation functions, depending on the specific task. Here, $E_i(x)$ denotes the output of expert $i$, $x$ represents an input to the model, and $N$ is the number of experts~\cite{shazeer2017outrageously}. Gating weights $g_i(x)$, detailed in Equation~\ref{eq:gating_weights_MoE}, are computed using a softmax function, with $a_i(x)$ representing the activation for an expert $i$ given the input $x$. The gating network uses the input data to determine which expert is best suited for the task. 

%\vspace{-1.5ex}

\begin{equation}
y(x)=\sum_{i=1}^N g_i(x) \cdot E_i(x)
\label{eq:MoE_output}
\end{equation}

%\vspace{-1.0ex}

\begin{equation}
g_i(x)=\frac{\exp \left(a_i(x)\right)}{\sum_{j=1}^N \exp \left(a_j(x)\right)}, \quad i=1,2, \ldots, N
\label{eq:gating_weights_MoE}
\end{equation}

%\vspace{-1.5ex}
% Not that we need another section, but now it feels remiss to not mention model merging, which has recently become very popular in the LLM space ...
% https://huggingface.co/collections/osanseviero/model-merging-65097893623330a3a51ead66
% https://arxiv.org/abs/2401.10491
% https://slgero.medium.com/merge-large-language-models-29897aeb1d1a
% https://huggingface.co/blog/mlabonne/merge-models

\subsection{Model Merging}
Model merging is a technique used to combine the parameters of multiple task-specific pre-trained LLMs to create a new and improved language model~\cite{yadav2023resolving}. Initially, this involves the process of selecting base models and aligning the architectures of chosen models to ensure compatibility. Techniques such as parameter averaging~\cite{matena2022merging} or knowledge distillation~\cite{khanuja2021mergedistill, hinton2015distilling} are then employed to integrate the knowledge from these models. Additionally, various algorithms, including task vector arithmetic~\cite{ilharco2022editing}, TIES~\cite{yadav2023resolving}, and DARE~\cite{yu2023language} can be used for parameter merging, each with its own advantages and considerations, such as computational complexity and the ability to handle models trained on different tasks. Following integration, the merged model undergoes fine-tuning on task-specific data to refine its representations and potentially optimize overall performance. The resulting merged model retains the knowledge and capabilities of its constituent models, leading to enhanced performance and capabilities across tasks compared to traditional methods of training a single model from scratch, as well as improved robustness and resource efficiency~\cite{wan2024knowledge}. However, challenges such as ensuring compatibility between models, managing computational complexity, and avoiding performance degradation must be addressed~\cite{wan2024knowledge, ainsworth2022git}.

\subsection{Diffusion Models}

Diffusion models are specifically designed for generating images and data samples~\cite{kingma2021variational}. These models are trained to generate realistic samples by modeling the diffusion process of a data distribution. Different approaches like \ac{NCE}~\cite{zach2023fully} and score-based generative modeling~\cite{song2020score} exist within the domain of diffusion models in Generative \ac{AI}. They operate by iteratively adding noise to a given initial image and subsequently learning to reverse this process to generate new, realistic, and high-quality images of varying styles and complexities~\cite{dhariwal2021diffusion, song2019generative}. As shown in Equation~\ref{eq:diffusion_models}, the general idea is to model the data distribution as a diffusion process, where the data is transformed from a simple distribution to the target distribution through a series of steps. Here, $x_t$ represents the data at time step $t$, $f$ denotes a diffusion process that transforms the data from $x_{t-1}$ to $x_t$, $\theta_t$ represents the parameters of the diffusion process at time step $t$, and $\epsilon_t$ represents a sample from a noise distribution $t$. This approach has led to the development of generative models such as \ac{DSM} and diffusion probabilistic models. The underlying idea is to transform a simple distribution through a series of steps to match the target distribution of real data. The generative process involves reversing these steps to generate new samples. Diffusion-based generative models, such as DALL-E 2~\cite{ramesh2022hierarchical, ramesh2021zero}, Imagen~\cite{saharia2022photorealistic}, stable diffusion~\cite{rombach2022high}, and others, are a class of probabilistic models that describe the evolution of an image from a simple initial distribution to the desired complex distribution~\cite{sohl2015deep}. 

\begin{equation}
x_t=f(x_{t-1}, \theta_t,\epsilon_t)
\label{eq:diffusion_models}
\end{equation}

\vspace{1.0ex}
\noindent \textbf{Stable Diffusion}. Text-to-image generation involves creating visual content based on textual descriptions~\cite{reed2016generative}. Stable diffusion is an open-source text-to-image diffusion model that generates diverse and high-quality images based on textual prompts\footnote{\url{https://stablediffusionweb.com/}}. This model operates by taking a noisy image as input and gradually denoising it to generate the desired output. The denoising process is guided by a text prompt, providing information about the desired content and style of the image.

\vspace{1.0ex}
\noindent \textbf{Midjourney}. Midjourney is a text-to-image diffusion model that, like Stable Diffusion~\cite{rombach2022high}, leverages prompts to generate unique and artistic images~\cite{midjourney2022}. However, it is a closed-source Generative \ac{AI} project requiring a paid subscription. This setup consequently may discourage community collaboration and development, leaving some users with less control over the underlying model compared to open-source alternatives like stable diffusion~\cite{rombach2022high}.

\subsection{Multimodal Generative Models}
Multimodal generative models represent a significant advancement in AI. These models possess the capability to understand and create content by leveraging various data types, such as text, images, and audio~\cite{wu2018multimodal, suzuki2022survey}. This integration of different data modalities enables these models to capture a more comprehensive understanding of concepts~\cite{shi2019variational}. By utilizing information from these diverse sources, multimodal generative models aim to overcome the limitations inherent in traditional models that focus solely on a single data type~\cite{suzuki2022survey}. Unimodal methods, traditional approaches that primarily focus on a single modality, such as text or images, have limitations in capturing the full complexity of real-world data~\cite{suzuki2022survey}. For example, text-based models may lack the ability to incorporate visual or emotional context into their understanding, while image-based models might lack textual or semantic understanding~\cite{suzuki2022survey}. Multimodal generative models address these limitations by integrating information from different modalities, such as text, images, and audio. This allows them to achieve a better understanding of the data and subsequently generate content that reflects the richness of human expression and experience. However, training multimodal models comes with its own set of challenges. These models can be computationally expensive to train and require large amounts of labeled data for each modality~\cite{suzuki2022survey}. Additionally, finding effective techniques to seamlessly integrate information from different modalities remains an active area of research~\cite{ngiam2011multimodal}. There are two main architectures used for multimodal learning: early fusion and late fusion~\cite{baltruvsaitis2018multimodal}. Early fusion combines data from all modalities at the beginning of the model, while late fusion processes each modality independently before combining the results. The ability of multimodal generative models to understand and create content across different data types makes them invaluable for a wide range of tasks requiring a deep understanding of multimodal data~\cite{radford2021learning}. Some real-world applications include generating realistic product descriptions with images for e-commerce platforms or creating personalized music recommendations based on a user's audio preferences and listening history. In addition to this, these models have demonstrated remarkable capabilities in various tasks, including medical imaging analysis, image captioning, text-to-image synthesis, video understanding, and audio-visual storytelling~\cite{radford2021learning}. By overcoming the limitations of unimodal models and offering new possibilities for creative content generation, multimodal generative models will play a significant role in the future of \ac{AI}.

\subsection{Applications of Generative AI}
Generative AI models are powerful tools for understanding and generating data with applications in various domains, including the following.

\vspace{1.0ex}
\noindent \textbf{Image Generation and Analysis}. Advanced Generative \ac{AI} models have demonstrated remarkable capabilities in generating high-quality images, such as photorealistic faces and scenes~\cite{radford2018improving}. Generative AI models have been employed in developing complex systems capable of generating and understanding multimodal data such as text and images. For example, the work in~\cite{yu2022scaling} proposes a large-scale autoregressive model that generates high-quality and content-rich images from text descriptions. Additionally, DALL-E is a generative model introduced by Ramesh \etal~\cite{ramesh2022hierarchical, ramesh2021zero}, which produces images from textual descriptions. Unlike traditional image generation models that rely on pixel-level manipulations or predefined templates, DALL-E operates at a semantic level, understanding textual prompts and synthesizing corresponding images. The work in~\cite{karras2019style} introduces a novel architecture specifically designed for generating high-quality facial images. This architecture utilizes a style-based generator, demonstrating advancements in synthesizing diverse and realistic images. Furthermore, Generative AI models can also be employed in image-to-image translation~\cite{park2019semantic}, which involves converting images from one domain to another, such as enabling the conversion of satellite images into maps or black-and-white photos into color. The work by Zhu \etal~\cite{zhu2017unpaired} presents a model designed for unpaired image-to-image translation. This model utilizes cycle-consistent adversarial networks to learn mappings between two image domains without requiring paired training examples, making it versatile for various applications~\cite{zhu2017unpaired}. Unlike DALL-E~\cite{ramesh2021zero}, which primarily focuses on generating images, \ac{CLIP} learns to understand the relationships between text and images in a paired manner~\cite{radford2021learning}. Through contrastive learning, \ac{CLIP} pre-trains on vast amounts of image-text pairs, enabling it to encode both modalities into a shared embedding space~\cite{radford2021learning}. CLIP's cross-modal understanding enables a wide range of applications beyond traditional image analysis tasks. By associating images with their textual descriptions, \ac{CLIP} can perform tasks such as image classification, object detection, and even zero-shot learning, where it recognizes objects or concepts not seen during training~\cite{radford2021learning}. \ac{CLIP} is built upon a dual-encoder architecture, featuring separate encoders for processing images and text. This architectural design allows \ac{CLIP} to independently encode visual and textual inputs into distinct feature spaces, facilitating effective cross-modal understanding. For image processing, \ac{CLIP} often employs \ac{CNNs} or \ac{ViT} to extract visual features~\cite{dosovitskiy2020image}. The image encoder within CLIP processes visual inputs, such as images, using \ac{CNNs}. Through pre-training on large-scale image datasets, the image encoder learns to extract hierarchical visual features that capture important characteristics of the input images. These features are then encoded into a high-dimensional representation space. On the other hand, the text encoder in CLIP processes textual inputs, such as captions or descriptions, using transformer architectures~\cite{vaswani2017attention, devlin2019bert}. Transformers are capable of modeling sequential data like text, allowing the text encoder to capture semantic information and contextual relationships within textual inputs. Through pre-training on large-scale text corpora, the text encoder learns to encode textual inputs into a corresponding feature space. Despite having separate encoders for images and text, CLIP achieves cross-modal understanding by mapping both image and text embeddings into a shared embedding space. This shared space facilitates direct comparisons between visual and textual representations, enabling CLIP to determine the semantic similarity between them~\cite{radford2021learning}. During pre-training, \ac{CLIP} leverages contrastive learning objectives to align similar pairs of image-text embeddings while maximizing the distance between dissimilar pairs, thereby enhancing its ability to understand and relate visual and textual inputs effectively~\cite{radford2021learning}.

\vspace{1.0ex}
\noindent \textbf{Video Generation}. Advanced Generative \ac{AI} models have not only demonstrated remarkable capabilities in generating high-quality images but have also begun to tackle the challenge of video generation. Recent advancements in AI, such as Sora developed by OpenAI~\cite{openaisora, brooks2024videoworldsimulators}, have enabled the generation of realistic and dynamic video content from textual descriptions. Similar to its image counterpart DALL-E~\cite{ramesh2022hierarchical}, Sora operates at a semantic level, understanding textual prompts and synthesizing corresponding video sequences~\cite{openaisora, brooks2024videoworldsimulators}. Video generation involves creating coherent and visually appealing sequences of frames that align with the provided textual instructions~\cite{brooks2024videoworldsimulators}. These models typically employ architectures designed to capture temporal dependencies (\ie relationships between frames over time) and spatial relationships (\ie relationships between objects within a single frame). By understanding the semantic context of the text, these models generate videos that accurately reflect described scenes while exhibiting smooth transitions and realistic motion. In addition to video generation, as explained above, AI models are capable of multimodal generation, where textual prompts can result in the synthesis of both images and videos. This capability enhances the quality of generated content, enabling diverse applications in storytelling, content creation, and multimedia production. Video generation has the potential to revolutionize various domains, including the entertainment industry, education and training, augmented reality and virtual reality applications, automation of video editing tasks, and \etc

\vspace{1.0ex}
\noindent \textbf{Text Generation}. Advances in Generative AI models can generate human-quality text, including translations, and responses to natural language questions~\cite{brown2020language, manyika2023overview}. Text generation models learn patterns and relationships in language from vast amounts of text data~\cite{brown2020language, manyika2023overview}.

\vspace{1.0ex}
\noindent \textbf{Code Generation}. Widely adopted \ac{AI} tools utilize generative \ac{AI} techniques to analyze the context of the code being written and suggest relevant code completions that can significantly improve programmers' and engineers' productivity by reducing the time spent manually typing codes~\cite{chen2021evaluating}.

\vspace{1.0ex}
\noindent \textbf{Drug Discovery}. Generative \ac{AI} models are increasingly being utilized in various aspects of drug discovery, providing innovative approaches to developing new drugs and accelerating the identification and design of novel therapeutic agents~\cite{zeng2022deep, altae2017low}. Furthermore, Generative \ac{AI} models have demonstrated the capability to identify new applications for drug repurposing~\cite{aliper2016deep}.

\vspace{1.0ex}
\noindent \textbf{Material Discovery}. Advanced \ac{ML} and \ac{DL} techniques, particularly Generative models, are being employed to explore and predict novel materials with desirable properties~\cite{merchant2023scaling}. The application of Generative AI models in material science~\cite{gomes2019artificial} can significantly accelerate the material discovery process by guiding experimental efforts, predicting new materials, and optimizing existing materials~\cite{pyzer2022accelerating}.

\vspace{1.0ex}
\noindent \textbf{Fraud Detection}. Generative \ac{AI} models have proven effective in detecting fraud by identifying patterns indicative of fraudulent activity~\cite{langevin2022generative}. Furthermore, these models can also be employed in identifying anomalies in data~\cite{schlegl2017unsupervised, schlegl2019f}. 

\vspace{1.0ex}
\noindent \textbf{Personalization}. Generative \ac{AI} models can be used in personalization to tailor content, recommendations, or user experiences based on individual preferences~\cite{chen2019generative, adiwardana2020towards}. This customization can involve generating personalized recommendations or creating personalized user experiences. For example, Netflix uses Generative AI to recommend movies and TV shows to its users, while Spotify leverages Generative \ac{AI} to create custom playlists.

\begin{comment}

\begin{table*}[!t]
\caption{Summary of the differences between the traditional and modern approaches to language modeling.}
\label{language_models_differences}
\centering
\begin{tabular}{l|c|c|c}
%\hline
\toprule
\rowcolor[HTML]{EFEFEF} 
\textbf{Feature}      & \textbf{Statistical Language Models} & \textbf{Neural Network Language Models} & \textbf{Transformer Language Models} \\ \midrule %\hline
Dataset size     & Small to medium & Medium to large & Massive \\ 
Training method       & Statistical methods                  & Neural networks                         & Neural networks                      \\ %\hline
Model architectures & N-gram models & RNNs and transformers & Transformers\\
Pre-training & Not typically pre-trained & Often pre-trained on large corpora of text & Commonly pre-trained on massive datasets \\
Accuracy (performance)             & Good                                 & Excellent                               & Excellent                            \\ %\hline
Computational cost    & Low to medium                        & High                                    & Very high                            \\ %\hline
Sensitivity to errors & High                                 & Medium                                  & Low                                  \\ %\hline
Overfitting risk  & Low  & Medium      & High        \\ \bottomrule 
\end{tabular}
\end{table*}

\end{comment}

\section{Language Modeling}
\label{language_modeling}

The use of language models is pervasive in various modern \ac{NLP} applications. In these models, the probability of different sequences of words is often modeled as the product of local probabilities, as expressed in Equation~\ref{equation:language_model_probability}, where $w_i$ represents the $i^{th}$ word in the sequence, and $h_i$ represents the word history preceding $w_i$. The formulation in Equation~\ref{equation:language_model_probability} summarizes the conditional dependencies between words in a sequence, allowing language models to capture complex linguistic patterns. Leveraging such models has proven instrumental in tasks ranging from machine translation and speech recognition to text generation and sentiment analysis~\cite{kalchbrenner2013recurrent, papineni2002bleu}.

\begin{equation}
P\left(w_1, w_2, \ldots, w_n\right)=\prod_{i=1}^n P\left(w_i \mid h_i\right)
\label{equation:language_model_probability}
\end{equation}

The following are some of the main approaches to traditional and modern approaches to language modeling. %Table~\ref{language_models_differences} summarizes the main differences between these language models.

\subsection{Statistical Language Models}

Statistical language models are based on the idea that the probability of a word appearing in a sentence is related to the probability of the words that came before it~\cite{liu2005statistical}. These models are trained on large corpora of text, and they use statistical methods to learn the probabilities of different sequences of words. Such models, including \emph{n-gram} models and models based on maximum entropy, often use conditional probability to estimate the likelihood of a word given its context~\cite{roark2007discriminative, khudanpur2000maximum}. Equation~\ref{statistical_language_models} is derived from the maximum likelihood estimation, where the probability of a word given its context is estimated by the ratio of the count of the specific context-word pair to the count of the context alone. In Equation~\ref{statistical_language_models}, $P\left(w_1, w_2, \ldots, w_n\right)$ denotes the conditional probability of the word, given the preceding word $w_{n-1}$, $C\left(w_{n-1}, w_n\right)$ represents is the count of occurrences of the bigram (word $w_{n-1}$, word $w_n$) in the training data, and the $C\left(w_{n-1}\right)$ represents the count of occurrences of the word $w_{n-1}$ in the training data. For higher-order \emph{n-gram} models, the equation is extended to consider a longer history of words as shown in Equation~\ref{n-gram_higher_order}.

\begin{equation}
P\left(w_n \mid w_{n-1}\right)=\frac{C\left(w_{n-1}, w_n\right)}{C\left(w_{n-1}\right)}
\label{statistical_language_models}
\end{equation}

%\vspace{-1.5ex}
\begin{multline}
%\begin{equation}
P\left(w_n \mid w_{n-1}, w_{n-2}, \ldots, w_1\right)=\\
\frac{C\left(w_{n-1}, w_{n-2}, \ldots, w_1, w_n\right)}{C\left(w_{n-1}, w_{n-2}, \ldots, w_1\right)}
\label{n-gram_higher_order}
%\end{equation}
\end{multline}

\subsection{Neural Network Language Models}

Neural network language models, particularly those based on \ac{RNNs} or transformer architectures, model the probability of a word given its context using a neural network. Actual neural network language models can have variations based on the specific architecture used (\eg recurrent or transformer-based). However, the simplified representation of such models can be broken down into the hidden state calculation and \emph{softmax} calculation as shown in Equations~\ref{hiden_calculation} and~\ref{softmax_calculation} respectively. Equation~\ref{hiden_calculation} shows the hidden state calculation where $\mathbf{h}_{n-1}$ denotes the hidden state of the neural network at time step $n-1$, $\mathbf{W}_{h}$ denotes the weight matrix for the hidden state transition, $\mathbf{U}_{h}$ shows the weight matrix for the word embedding transition, $\mathbf{E}_{n-2}$ denotes the embedding vector of the word $w_{n-2}$, and $tanh$ is the hyperbolic tangent activation function. Equation~\ref{softmax_calculation} shows the \emph{softmax} output calculation which computes the conditional probability distribution over the vocabulary for the next word $w_n$ where $P\left(w_n \mid w_{n-1}, w_{n-2}, \ldots, w_1\right)$ denotes the Conditional probability of the word $w_n$ given the history $w_{n-1}, w_{n-2}, \ldots, w_1$, $\mathbf{W}_o$ shows the weight matrix for the output layer, $\mathbf{h}_{n_1}$ is the hidden state of the neural network at time step $n-1$, where the $softmax$ is the \emph{softmax} function, converting the network's output into probabilities.

\begin{equation}
\mathbf{h}_{n-1}=\tanh \left(\mathbf{W}_h \cdot \mathbf{h}_{n-2}+\mathbf{U}_h \cdot \mathbf{E}_{n-2}\right)
\label{hiden_calculation}
\end{equation}

\begin{multline}
%\begin{equation}
P\left(w_n \mid w_{n-1}, w_{n-2}, \ldots, w_1\right)=\\
\operatorname{softmax}\left(\mathbf{W}_o \cdot \tanh \left(\mathbf{h}_{n-1}\right)\right)
\label{softmax_calculation}
%\end{equation}
\end{multline}

\subsection{Transformer Language Models}
\label{large_language_models}

Transformer language models are based on the idea of attention, which allows the model to focus on the most relevant parts of the input sequence when making predictions~\cite{vaswani2017attention, devlin2019bert, brown2020language}. Such models leverage pre-training to achieve strong performance across various \ac{NLP} tasks. According to~\cite{vaswani2017attention}, the transformer architecture offers several advantages over traditional recurrent or convolutional neural networks. It enables significantly more parallelization for faster training, achieves state-of-the-art results in machine translation with shorter training times, reduces the complexity of relating distant input positions, and effectively models long-range dependencies while handling variable-length sequences~\cite{vaswani2017attention}. The transformer model achieves state-of-the-art results in machine translation by employing attention mechanisms, enabling it to capture long-range dependencies and process variable-length sequences without padding or truncation~\cite{vaswani2017attention}. Moreover, it simplifies the computation of relationships between distant positions, leading to enhanced parallelization, faster training, and superior performance compared to traditional neural networks.

\vspace{1.0ex}
\noindent \textbf{Self-Attention Mechanism}. The Transformer architecture revolutionized sequence modeling by introducing a self-attention mechanism, eliminating the need for recurrent or convolutional structures. The self-attention mechanism essentially computes a weighted sum of input representations, where each position in the input sequence is allowed to attend to all other positions with different weights. This mechanism allows the model to capture long-range dependencies between distant words in a sentence, which is important for tasks such as machine translation and text summarization. Given an input sequence $X=\left\{x_1, x_2, \ldots, x_n\right\}$, the self-attention mechanism computes the output vector $Y=\left\{y_1, y_2, \ldots, y_n\right\}$. As shown in Equation~\ref{eq:self-attention}, the attention mechanism computes a set of attention scores, which are then used to calculate a weighted sum of the input vectors. Here, $Q_i$, $K_j$, $v_j$ are the query, key, and value vectors for the $i^th$ output element and $j^th$ input element, respectively, and $d_k$ is the dimension of the key vectors~\cite{vaswani2017attention}. The attention score $a_{i j}$ for the $i^{th}$ element in the output sequence and the $j^{th}$ element in the input sequence is computed as shown in Equation~\ref{Eq:attention_score}. Here, $e_{i j}$, commonly represented as $Q^T_i \cdot K_j $, is the attention energy or compatibility function between the $i^{th}$ element in the output sequence and the $j^{th}$ element in the input sequence. Once the attention scores are computed, the weighted sum of the input vectors is calculated to obtain the context vector for each output element as shown in Equation~\ref{Eq:context_vector} where $V_j$ is the value vector for the $j^{th}$ input element.

\begin{equation}
    %$$
y_i=\sum_{j=1}^n \frac{\exp \left(e_{i j}\right)}{\sum_{k=1}^n \exp \left(e_{i k}\right)} \cdot v_j
,
\text{~where}~
%$$
e_{i j}=\frac{\left(Q_i \cdot K_j\right)}{\sqrt{d_k}}
\label{eq:self-attention}
\end{equation}

\begin{equation}
a_{i j}=\frac{\exp \left(e_{i j}\right)}{\sum_{k=1}^n \exp \left(e_{i k}\right)}
\label{Eq:attention_score}
\end{equation}

\begin{equation}
c_i=\sum_{j=1}^n a_{i j} \cdot V_j
\label{Eq:context_vector}
\end{equation}

\vspace{1.0ex}
\noindent \textbf{Multi-Head Self-Attention}. The multi-head self-attention mechanism is a variant of the self-attention mechanism that introduces multiple attention heads to capture different aspects of the relationships in the input sequence~\cite{vaswani2017attention}. The transformer model uses multiple self-attention heads in parallel across multiple heads to capture different aspects of the relationships within the input sequence instead of performing a single attention function with $d_{model}$-dimensional keys, value vectors, and queries. This allows the model to learn more complex representations of the input, which can improve performance on a variety of \ac{NLP} tasks. As shown in Equation~\ref{eq:multi-head-self-attention}, the outputs of these heads are concatenated and linearly transformed~\cite{vaswani2017attention} where the transformations are parameter matrices $W_i^Q \in \mathbb{R}^{d_{\text {model }} \times d_k}, W_i^K \in \mathbb{R}^{d_{\text {model }} \times d_k}, W_i^V \in \mathbb{R}^{d_{\text {model }} \times d_v}$ and $W^O \in \mathbb{R}^{h d_v \times d_{\text {model }}}$. Here, $W_i^Q$, $W_i^K$, $W_i^V$, and $W^O$ are learned weight matrices. This allows the model to learn a wider range of relationships between words in the input sequence.

\begin{multline}
    \text{MultiHead}(Q, K, V)= \text{Concat}\left({head}_1, \ldots, {head}_{\mathrm{h}}\right).W^O \\
    \text{Where}~\\ 
    \text{head}{_i} = \text{SelfAttention}\left(Q W_i^Q, K W_i^K, V W_i^V\right) 
\label{eq:multi-head-self-attention}
\end{multline}

\vspace{1.0ex}
\noindent \textbf{\ac{FFN}}. The \ac{FFN} is an important component of the transformer architecture. It is responsible for processing information from the self-attention mechanism across all positions in the input sequence~\cite{vaswani2017attention}. The \ac{FFN} consists of two fully connected linear transformations with a \ac{ReLU} activation function in between them. This structure allows the \ac{FFN} to learn complex non-linear relationships between the input features~\cite{vaswani2017attention}. The \ac{FFN} is applied independently to each position in the input sequence, ensuring that each position can interact with all other positions~\cite{vaswani2017attention}. This parallelized approach makes the \ac{FFN} computationally efficient and scalable to long input sequences. The output of the self-attention mechanism is then passed through a position-wise feed-forward network as shown in Equation~\ref{eq:ffn}, where the learned parameters $W_1$ and $W_2$ are learned weight matrices while $b_1$ and $b_2$ are the learned bias vectors. As shown in Equation~\ref{eq:SwishGELU}, other works have proposed replacing the \ac{ReLU} activation function with other nonlinear activation functions such as \ac{GELU} $(x)=x \Phi(x)$~\cite{hendrycks2016gaussian} where $\Phi(x)$ is the standard Gaussian cumulative distribution function, and $\operatorname{Swish}_\beta(x)=x \sigma(\beta x)$~\cite{ramachandran2017searching}.

\begin{equation}
    \text{FFN($x$)}=max(0, x.W_1+b_1). W_2 + b_2
    \label{eq:ffn}
\end{equation}

\begin{equation}
\begin{aligned}
FFN_{\text{GELU }}\left(x, W_1, W_2\right)={\text{GELU}}\left(x W_1\right) W_2 \\
FFN_{\text{Swish}}\left(x, W_1, W_2\right) ={\text{Swish}}_1\left(x W_1\right) W_2
\end{aligned}
\label{eq:SwishGELU}
\end{equation}

\begin{table*}[t!]
\caption{A list of some of the state-of-the-art \ac{LLMs} well-suited for a wide range of \ac{NLP} tasks.}
\label{table:LLMs_comparison}

 \begin{minipage}{\textwidth}
  \centering
\begin{tabular}{c|c|c|c|c|c}
\toprule
\rowcolor[HTML]{EFEFEF} 
\textbf{Year of Release} & \textbf{LLMs}        & \textbf{\begin{tabular}[c]{@{}c@{}}Number of \\ Parameters\end{tabular}} & \textbf{Number of Training Tokens} & \textbf{\begin{tabular}[c]{@{}c@{}}Learning Rate \\ (\emph{Default})\end{tabular}} &\textbf{Developer} \\ \midrule %\hline
2017          & Transformer~\cite{vaswani2017attention} & 530 Million & Not explicitly stated & 1x$10^{-3}$& Google AI \\ \midrule 
2018          & BERT~\cite{devlin2019bert}        & 340 Million                                                                       & 250 Billion & 5x$10^{-5}$ & Google AI          \\ \midrule 
2019         & GPT-2~\cite{radford2019language}               & 1.5 Billion                                                                      & 40 Billion & 1x$10^{-5}$& OpenAI             \\ \midrule 
2020        & T5~\cite{raffel2020exploring}   & 11 Billion    & 1 Trillion & 5x$10^{-5}$ & Google AI \\ \midrule
2020        & GPT-3~\cite{brown2020language}  & 175 Billion    & 300 Billion & 6x$10^{-5}$& OpenAI  \\ \midrule 
2020        & Gopher~\cite{rae2021scaling}    & 280 Billion   & 300 Billion &  4x$10^{-5}$ & Google AI \\ \midrule
2021         & Jurassic-1 Jumbo~\cite{lieber2021jurassic}    & 178 Billion                                                                      & 300 Billion & 6x$10^{-5}$& AI21 Labs          \\ \midrule
2021          & Megatron-Turing NLG~\cite{smith2022using} & 530 Billion                                                                      & 270 Billion & 5x$10^{-5}$ & NVIDIA             \\ \midrule
2022        & Chinchilla~\cite{hoffmann2022training}  & 70 Billion      & 1.4 Trillion & 1.25x$10^{-4}$& Deep Mind \\ \midrule
2022        & LaMDA~\cite{thoppilan2022lamda} & 137 Billion & 768 Billion & Not explicitly stated & Google AI \\ \midrule
2022        & GPT-3.5 (InstructGPT)~\cite{ouyang2022training} & 175 Billion & Not explicitly stated\footnote{OpenAI has not officially stated the exact number of training tokens used for GPT-3.5 (InstructGPT) until the publication of this work. However, it is rumored to be in the range of 600 to 700 billion tokens.} & Not explicitly stated & OpenAI \\ \midrule
2022        & GPT-3.5 (ChatGPT)  & 175 Billion\footnote{OpenAI has not officially stated the size of GPT-3.5 until the publication of this work. However, it is rumored to be 175 Billion parameters.} & Not explicitly stated\footnote{OpenAI has not officially stated the exact number of training tokens used for GPT-3.5 (ChatGPT) until the publication of this work. However, it is rumored to be in the range of 600 to 700 billion tokens.} & 5x$10^{-5}$ & OpenAI \\ \midrule
2022          & PaLM~\cite{narang2022pathways, chowdhery2022palm}    & 540 Billion                                & 780 Billion & Not explicitly stated & Google AI  \\ \midrule 
2023   & LLaMA~\cite{touvron2023llama2}  &  65 Billion & 1.4 Trillion & 1.5x$10^{-4}$ & Meta AI \\ \midrule 
2023          & Llama 2~\cite{touvron2023llama2} & 70 Billion & 2 Trillion & 1.5x$10^{-4}$& Meta AI \\ \midrule
2023          & PaLM 2~\cite{anil2023palm} & 340 Billion\footnote{Google has not officially stated the size of PaLM 2 until the publication of this work. However, it is rumored to be 340 Billion parameters.} & 3.6 Trillion & Not explicitly stated & Google AI \\ \midrule
2023          & GPT-4~\cite{openai2023gpt4} & 1-1.76 Trillion\footnote{OpenAI has not officially stated the size of GPT-4 until the publication of this work. However, it is rumored to be between 1 and 1.76 Trillion parameters.}        &Not explicitly stated\footnote{OpenAI has not officially stated the exact number of training tokens used for GPT-4 until the publication of this work. However, it is rumored to be in the range of 10 to 100 Trillion tokens.} & Not explicitly stated & OpenAI             \\ \midrule 
2023          & Gemini~\cite{google2023gemini} & Not explicitly stated\footnote{Google has not officially stated the size of Gemini Pro or Ultra until the publication of this work. However, Nano has two versions at 1.8 Billion and 3.25 Billion parameters.} & Not explicitly stated\footnote{Google has not officially stated the exact number of training tokens for any of the Gemini models until the publication of this work. However, they do follow the approach of \cite{hoffmann2022training}.} & Not explicitly stated & Google AI \\

\bottomrule

\end{tabular}
\end{minipage}
\end{table*}

\begin{figure*}[h!]
%\vskip-0.45cm
	\centering
	\includegraphics[width= \linewidth]{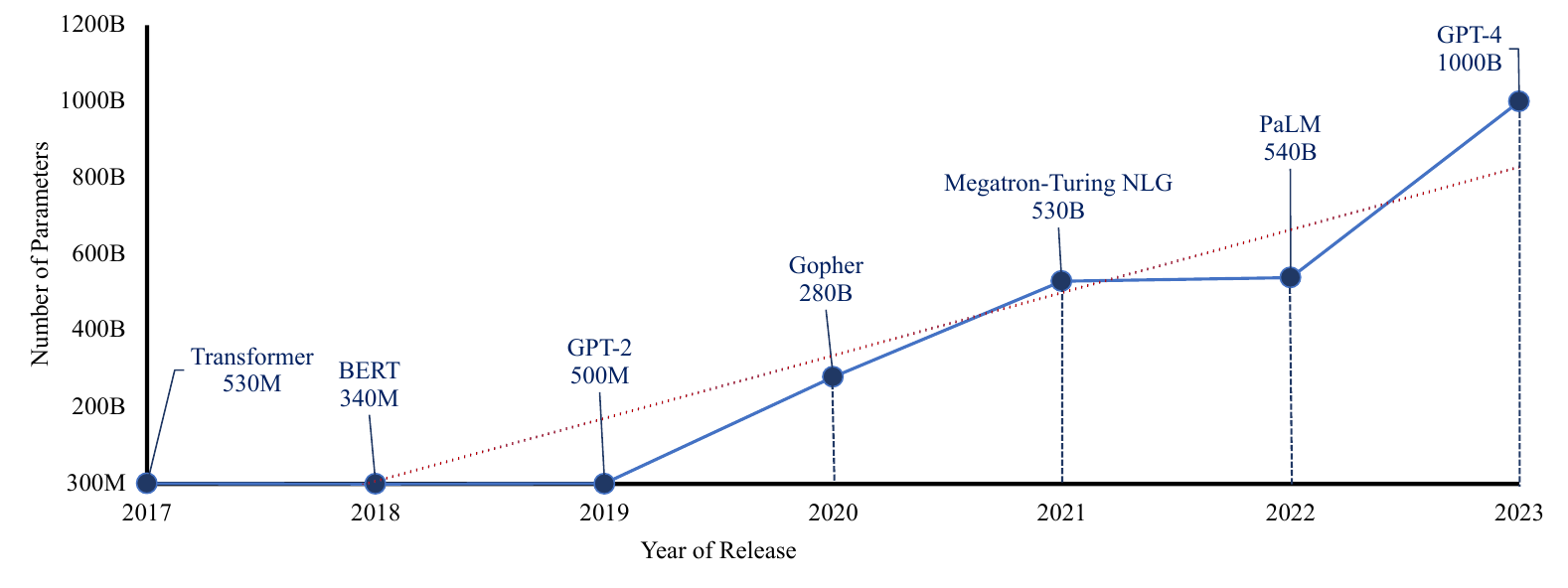}
	\caption{Timeline and model size of \ac{LLMs} (\emph{M=millions, B=billions}).}
	\label{Fig:time-line-LLMs}
 %\vskip-0.45cm
\end{figure*}

\begin{table}[t!]
\caption{A performance comparison of some of the state-of-the-art \ac{LLMs} well-suited for a wide range of \ac{NLP} tasks, as reported on PapersWithCode. (\emph{MMLU = Massive Multitask Language Understanding, GSM8K = Grade School Math, ARC = Abstraction and Reasoning Corpus})}
\label{table:LLMs_performance}

\begin{minipage}{\textwidth}
%\centering
\begin{tabular}{c|c|c|c}
\toprule
\rowcolor[HTML]{EFEFEF}

\textbf{Model}    & \textbf{MMLU} & \textbf{GSM8K} & \textbf{ARC (Challenge)} \\ \midrule
Gemini Pro        & 79.1          & 86.5           & --   \\ \midrule
Gemini Ultra      & \textbf{90.0} & \textbf{94.4}  & --   \\ \midrule
GPT-3             & 53.9          & 55.0           & 53.2 \\ \midrule
GPT-3.5 (ChatGPT) & 70.0          & 57.1           & 85.2 \\ \midrule
GPT-4             & 86.5          & 92.0           & \textbf{96.3} \\ \midrule
LLaMA (65B)       & 63.4          & 50.9           & 56.0 \\ \midrule
Llama 2 (70B)     & 68.9          & 56.8           & 67.3 \\ \midrule
PaLM (540B)       & 69.3          & 82.1           & 87.1 \\ \midrule
PaLM 2            & 81.2          & 91.0           & 95.1 \\

\bottomrule

\end{tabular}
\end{minipage}
\end{table}

In the context of language models, the transformer architecture facilitates the training of \ac{LLMs}, such as \ac{GPT}~\cite{goodfellow2014generative}. \ac{LLMs} are a type of generative AI model that is specifically trained on large corpora of text data. In recent years, \ac{LLMs} have emerged as transformative breakthroughs in the field of \ac{AI}, \ac{NLG}, and \ac{NLU}~\cite{mcshane2017natural} due to their remarkable capabilities in understanding and generating human-like text and other forms of content~\cite{radford2018improving}. \ac{LLMs} are trained on massive datasets comprising text and code, and they exhibit the ability to learn and perform a wide range of language tasks, including text generation, language translation~\cite{bahdanau2014neural}, text summarization~\cite{ouyang2022training}, sentiment analysis~\cite{nasukawa2003sentiment}, and question answering~\cite{di2019adapting}. These models are more powerful and versatile than traditional language models. \ac{LLMs} have revolutionized the way we interact with and leverage natural language data, and they are now used in a wide variety of applications, including chatbots~\cite{adiwardana2020towards}, machine translation systems~\cite{brants2007large, kalchbrenner2013recurrent}, and search engines. These models have experienced significant growth in terms of scale, complexity, and performance. Recently, several \ac{LLMs} have been introduced, with some of the largest dense language models that have scaled to billions of model sizes~\cite{smith2022using, rae2021scaling, lieber2021jurassic, brown2020language, thoppilan2022lamda}. These powerful models demonstrate the capability to perform a wide range of innovative \ac{NLP} tasks, including machine translation, text summarization, question answering, and code completion. To provide a comprehensive comparison of some well-known state-of-the-art LLMs, we have presented a list in Table~\ref{table:LLMs_comparison}. Figure~\ref{Fig:time-line-LLMs} shows a trend of some of the \ac{LLMs} and their corresponding number of parameters (model sizes). Some of the well-known state-of-the-art \ac{LLMs} include GPT~\cite{goodfellow2014generative} T5~\cite{raffel2020exploring}, Gopher~\cite{rae2021scaling}, LaMDA~\cite{touvron2023llama2}, \etc These models have demonstrated the power of pre-trained, massive neural networks for \ac{NLP} tasks. For example, \ac{GPT} can be used to generate realistic and coherent text, while \ac{BERT} can be used to extract complex meaning from text. Table~\ref{table:LLMs_performance} shows a performance comparison of some of the state-of-the-art \ac{LLMs} well-suited for a wide range of \ac{NLP} tasks, as reported on PapersWithCode\footnote{\url{https://paperswithcode.com/}}.

\subsection{Architecture of Transformer Models}
Transformer architectures have revolutionized \ac{NLP} tasks, such as sequence modeling, by effectively capturing long-range dependencies and modeling relationships between words. The advantages of the transformer architecture include enhanced parallelization, faster training, and the ability to model long-range dependencies efficiently. The attention mechanism allows the model to focus on relevant parts of the input sequence, contributing to its success in handling variable-length sequences without sacrificing performance. Recognizing the shift from encoder-decoder to decoder-only architectures, understanding pre-training strategies, and the advantages of transformer models provide a more nuanced perspective on their capabilities in Generative AI and various \ac{NLP} tasks. Here, we will distinguish between the original encoder-decoder architecture and the decoder-only architecture and the pre-training strategies of transformer models.

\vspace{1.0ex}

\noindent \textbf{Encoder-Decoder Architecture}. The encoder-decoder architecture serves as a fundamental structure in Transformer models, employed for sequence-to-sequence tasks such as machine translation, where an input sequence (source language) is transformed into an output sequence~\cite{vaswani2017attention, wu2023decoder}. In an encoder-decoder architecture, the model consists of two main components featuring multiple layers of self-attention and feedforward layers: an encoder and a decoder network. The encoder network processes the input sequence, capturing relevant information and creating a contextualized representation that encompasses semantic and syntactic details of the input.  Subsequently, the decoder network, in turn, utilizes this contextualized representation from the encoder to generate the output sequence step by step. At each step, the decoder attends to various parts of the encoder's output, facilitating the alignment of source and target language information. Both the encoder and decoder components typically employ the self-attention mechanism~\cite{vaswani2017attention}. This mechanism enables the model to weigh the importance of different positions in the input sequence during the generation of the output sequence, thereby allowing for the capture of long-range dependencies. Encoder-decoder architectures are commonly trained in a supervised and unsupervised manner, where the model is first pre-trained on a large corpus, then fine-tuned on provided with pairs of input sequences and corresponding target output sequences. The model learns to map input sequences to output sequences by minimizing a suitable loss function~\cite{vaswani2017attention}. However, the field has witnessed a significant shift with the emergence of decoder-only architectures, indicating a transition towards more flexible and potentially more powerful models.

\vspace{1.0ex}
\noindent \textbf{Decoder-Only Architecture}. The decoder-only architecture utilizes only the decoder component of the Transformer model~\cite{wu2023decoder, radford2018improving}. In this architecture, the model generates output sequences autoregressively, predicting one token at a time based on the preceding tokens without relying on an explicit encoder~\cite{wu2023decoder}. The absence of an encoder implies that the model does not receive direct information about the input sequence but instead uses its autoregressive nature to capture dependencies within the generated sequence itself. Decoder-only architectures leverage a specific variant of the self-attention mechanism. This mechanism allows the model to attend to different positions within the already generated sequence while predicting each new token, effectively capturing the necessary contextual information for generating coherent output~\cite{wu2023decoder}. These models are typically pre-trained on massive text corpora in an unsupervised manner~\cite{radford2018improving}. During this pre-training phase, the model learns general language representations, capturing both syntactic and semantic information~\cite{radford2018improving}. Subsequently, fine-tuning on specific tasks with labeled data allows the model to adapt to various downstream applications. One well-known example of the decoder-only architecture is the \ac{GPT}~\cite{brown2020language}. \ac{GPT} employs a stack of transformer decoder layers for autoregressive sequence generation~\cite{brown2020language}.

\subsection{Pre-training Strategies in Transformer Language Models}
One of the key factors behind the success of transformer-based language models is their pre-training on massive amounts of text data using self-supervised learning techniques~\cite{vaswani2017attention}. This pre-training stage equips the models with a robust understanding of language structure and semantics, enabling exceptional performance on various downstream \ac{NLP} tasks~\cite{devlin2019bert, radford2018improving}. Transformer language models, leveraging pre-training, have demonstrated outstanding performance across diverse \ac{NLP} tasks. In machine translation, the transformer's attention mechanism allows it to capture long-range dependencies, yielding state-of-the-art results without the need for excessive padding or truncation~\cite{vaswani2017attention}. Beyond translation, decoder-only architectures like \ac{GPT} have proven effective in tasks such as sentiment analysis, named entity recognition, and text completion~\cite{radford2018improving}.

\vspace{1.0ex}

\noindent \textbf{Self-Supervised Learning for Pre-training}. Unlike traditional supervised learning methods that demand extensive labeled data, self-supervised learning leverages the unlabeled nature of textual data. Common pre-training objectives in transformers involve tasks like predicting the next word in a sequence, also known as \ac{MLM}~\cite{devlin2019bert, yamaguchi2021frustratingly}, or reconstructing a sentence where certain words are replaced with special tokens (masked tokens)~\cite{devlin2019bert}. By tackling these tasks, the model learns contextual relationships between words and develops a strong understanding of grammatical structures. The pre-training phase serves as a critical foundation for downstream \ac{NLP} tasks. The learned representations from vast amounts of text data can be fine-tuned for specific tasks like sentiment analysis, question answering, or machine translation. This approach requires significantly less labeled data compared to training a model from scratch~\cite{howard2018universal}. Consequently, self-supervised learning not only improves the efficiency of \ac{NLP} model training but also enables them to perform effectively on tasks where obtaining large amounts of labeled data might be challenging.

\subsection{Long Sequence Language Models}

Long sequence language models are neural network architectures specifically designed to effectively handle long textual input sequences by leveraging the Transformer architecture~\cite{zaheer2020big}. While various architectures can handle longer sequences, Transformers are dominant due to their self-attention mechanisms, enabling parallel processing and capturing long-range dependencies, overcoming the sequential limitations of \ac{RNNs}. This, unlike traditional language models, enables long-sequence language models to efficiently capture long-range dependencies and relationships between words~\cite{zaheer2020big}.  Several long sequence language models address the limitations of standard Transformers by introducing modifications and additional features to their architectures.

\vspace{1.0ex}

\noindent \textbf{Transformer-XL}. Transformer-XL is an extension of the standard Transformer model designed to overcome the limitations of fixed-length contexts in traditional models~\cite{dai2019transformer}. It addresses the inherent limitation of the standard Transformer model, which employs a fixed-length context window, by employing two advanced mechanisms. These mechanisms enable the model to learn dependencies beyond a fixed length in language modeling and retain information from previous segments of the input sequence, thus enhancing its ability to process longer sequences more effectively~\cite{dai2019transformer}. The first mechanism, segment-level recurrence, allows the model to reuse hidden states from previous segments by propagating them through recurrent connections. This enables information flow across segments, facilitating the retention of context from previous segments and extending the context beyond a fixed length. Incorporating recurrence at the segment level empowers Transformer-XL to capture longer-term dependencies in the data~\cite{dai2019transformer}. In addition to the segment-level recurrence mechanism, Transformer-XL employs a novel relative positional encoding scheme~\cite{dai2019transformer}. This encoding scheme is crucial for enabling state reuse without causing temporal confusion, thereby allowing the model to effectively capture dependencies across longer sequences. By utilizing relative positional encodings instead of absolute ones, Transformer-XL ensures that information can be propagated across longer sequences without sacrificing temporal coherence. This encoding scheme plays a vital role in allowing the model to learn dependencies that extend beyond the fixed context length~\cite{dai2019transformer}. Furthermore, Transformer-XL incorporates a state reuse mechanism by caching a sequence of hidden states from previous segments, which can be reused during evaluation. As demonstrated in~\cite{dai2019transformer}, this state reuse mechanism significantly accelerates evaluation and enables the model to maintain context from earlier segments, contributing to its ability to capture long-term dependencies in sequences.

\vspace{1.0ex}

\noindent \textbf{XLNet Architecture}. XLNet represents a pre-training method for \ac{NLU} tasks~\cite{yang2019xlnet}. Building upon BERT's bidirectional context modeling~\cite{devlin2019bert}, XLNet addresses its limitations, such as the fixed-length context constraint. Unlike BERT, which relies on masked language modeling, XLNet achieves bidirectional context learning by maximizing the expected likelihood over all permutations of the factorization order~\cite{yang2019xlnet}. By utilizing an autoregressive formulation, XLNet ensures consistency between pretraining and fine-tuning stages, a limitation observed in BERT. As shown in Equation~\ref{xlnet_permutation}, instead of predicting the next word in a sentence given all previous words, XLNet predicts based on randomly chosen permutations of the input sequence. This approach encourages the model to consider all possible input permutations, effectively capturing the dependencies within the sequence. This involves randomly shuffling the order of elements in the sequence and then predicting each element based on its permuted context. In Equation~\ref{xlnet_permutation}, $x_1, x_2, \ldots, x_n$ represents the input sequence, $\pi$ is a random permutation of indices, and $P\left(x_{\pi(i)} \mid x_1, x_2, \ldots, x_{\pi(i-1)}\right)$ is the conditional probability of predicting the $i^{th}$ token given the previously predicted tokens and the current input sequence. During training, XLNet receives permuted sequences as input and predicts each element based on the surrounding elements in the shuffled order~\cite{yang2019xlnet}. This forces the model to learn contextual representations that are not dependent on the order of elements. Additionally, as shown in Equation~\ref{xlnet_generalized_autoregressive}, $x_1, x_2, \ldots, x_n$ denotes the the input sequence, and $P\left(x_i \mid x_1, x_2, \ldots, x_{i-1}\right)$ is the conditional probability of predicting the $i^{th}$ token given the previously predicted tokens $x_1, x_2, \ldots, x_n$. Equation~\ref{xlnet_generalized_autoregressive} represents the probability of generating the entire sequence $x_1, x_2, \ldots, x_n$ by factorizing it into conditional probabilities conditioned on the previously generated tokens.
XLNet incorporates a generalized autoregressive objective, similar to the one used in \ac{GPT} models, allowing diverse and coherent text generation. In Equation~\ref{xlnet_generalized_autoregressive}, Integrating ideas from Transformer-XL enhances XLNet's ability to handle long-range dependencies and capture contextual information efficiently~\cite{yang2019xlnet}. XLNet can be applied across a wide range of \ac{NLP} tasks including question answering, natural language inference, sentiment analysis, and document ranking~\cite{yang2019xlnet}. Its generalized autoregressive pretraining method enables effective handling of bidirectional contexts, long-range dependencies, and ensures consistency across pretraining and fine-tuning stages. Furthermore, XLNet's integration of Transformer-XL and advanced architectural designs improves performance on tasks involving longer text sequences and explicit reasoning.

\begin{equation}
P\left(x_1, x_2, \ldots, x_n\right)=\prod_{i=1}^n P\left(x_{\pi(i)} \mid x_1, x_2, \ldots, x_{\pi(i-1)}\right)
\label{xlnet_permutation}
\end{equation}

%\vspace{-1.5ex}
\begin{equation}
P\left(x_1, x_2, \ldots, x_n\right)=\prod_{i=1}^n P\left(x_i \mid x_1, x_2, \ldots, x_{i-1}\right)
\label{xlnet_generalized_autoregressive}
\end{equation}

\vspace{1.0ex}
\noindent \textbf{Longformer}. In the context of long sequence language models, the Longformer is a specialized architecture designed to improve the processing of long textual inputs~\cite{beltagy2020longformer}. It shares the transformer architecture's foundation but introduces modifications to the attention mechanism to accommodate the challenges posed by long sequences. It uses a locality-sensitive attention mechanism where each token attends only to its relevant local context and a few globally important tokens~\cite{zaheer2020big, beltagy2020longformer}. This attention only considers relevant subsequences around each token, improving efficiency for long sequences and it is adjusted as shown in Equation~\ref{longformer_attention} where $Q_i$, $K_j$, and $V_j$ are the query, key, and value vectors for positions $i$ and $j$ in the input sequence, respectively. The $mask\_matrix$ is used to mask certain positions, such as preventing attending to future positions during training or ignoring padding positions. In Equation~\ref{longformer_attention}, the division by $\sqrt{d_k}$ is a scaling factor that helps stabilize the gradients during training, where $d_k$ is the dimensionality of the key vectors. As described in~\cite{beltagy2020longformer, zaheer2020big}, Longformer's attention mechanism scales linearly with the sequence length, making it feasible to process long documents efficiently. It combines local windowed attention with task-motivated global attention. Local attention is primarily used to build contextual representations, while global attention allows Longformer to create full sequence representations for prediction~\cite{beltagy2020longformer}. In standard transformers, the self-attention mechanism considers interactions between all pairs of positions in the input sequence, leading to quadratic complexity.

\begin{equation}
\begin{aligned}
& \operatorname{Attention}\left(Q_i, K_j, V_j, \text {mask\_matrix}\right)= \\
& \operatorname{softmax}\left(\frac{Q_i K_j^T}{\sqrt{d_k}}+\text {mask\_matrix}_{i j}\right) \cdot V_j
\end{aligned}
\label{longformer_attention}
\end{equation}

\vspace{1.0ex}
\noindent \textbf{Sparse Transformers}. The standard transformer's attention mechanism calculates attention scores for all pairs of positions in a sequence, leading to quadratic time complexity~\cite{vaswani2017attention}. Sparse Transformers address this issue by considering only a subset of positions during attention computation~\cite{child2019generating}. This introduces sparsity, significantly reducing memory requirements and computational load, making them suitable for longer sequences~\cite{child2019generating}. As shown in Equation~\ref{Eq:sparse_transformers}, the sparse Transformer is a modified version of the standard attention mechanism used in transformers~\cite{child2019generating}. Sparse Transformer's attention, given a sequence of input embeddings $X$ with dimensions $N \times d$, where $N$ is the sequence length and $d$ the embedding dimension, the attention scores for position $i$ attending to position $j$ can be computed as shown in Equation~\ref{Eq:sparse_transformers}, where $Sp$ represents the sparse attention, $Q_i$ represents the query vector for position $i$, $K_j$ denotes the key vector for position $j$, ${Q_i K_j^T}$ represents the dot product of query and the key vectors, capturing the pairwise interactions between positions in the input sequence, $V_j$ represents the value vector for position $j$, and $M_{i j}$ denotes a binary mask element indicating whether vector position $i$ attends to vector position $j$. This demonstrates that the attention mechanism is computed for each pair of positions $i$ and $j$ based on their corresponding query, key, and value vectors. In global sparse attention, the mask $M_{i j}$ is generated by randomly selecting a fixed number of positions for each position $i$ to attend to. This introduces sparsity by limiting the attention to a small subset of positions in the sequence~\cite{child2019generating}. However, for local sparse attention, the mask $M_{i j}$ ensures that each position attends to a nearby local neighborhood. This reduces the computational complexity associated with attending to all positions and helps capture short-range dependencies efficiently~\cite{child2019generating}. The division by $\sqrt{d_k}$ serves as a scaling factor for numerical stability, where $d_k$ represents the dimensionality of the key vectors. Additionally, the binary mask $M_{i j}$ controls the sparsity pattern by allowing only certain positions to contribute to the attention scores. The $softmax$ function is applied to the masked and scaled dot product to normalize the scores and finally, the result is multiplied element-wise with the value matrix $V_j$. Some variations of Sparse Transformers incorporate adaptively determined sparsity based on the input sequence, task, or training phase~\cite{correia2019adaptively}, enhancing the model's flexibility and performance in handling diverse sequences~\cite{correia2019adaptively}.

\begin{equation}
\operatorname{Sp}\left(Q_i, K_j, V_j, M_{i j}\right)=\operatorname{softmax}\left(\frac{Q_i K_j^T \cdot M_{i j}}{\sqrt{d_k}}\right) \cdot V_j
\label{Eq:sparse_transformers}
\end{equation}

% For LLMs specifically, here's an article (with references) that takes about context window scaling techniques:
% https://towardsdatascience.com/why-and-how-to-achieve-longer-context-windows-for-llms-5f76f8656ea9
% These techniques should be mentioned here.

\subsection{Applications of LLMs}

\ac{LLMs} are a specific type of Generative \ac{AI} designed primarily for generating and understanding human language. In addition to the applications of Generative \ac{AI} explained in Section~\ref{generative_AI}, \ac{LLMs} can be employed for various other important tasks, such as the following.  %Some of the main applications of transformer-based models are summarized in Table~\ref{table:applications_of_transformer_models}.

\vspace{1.0ex}
\noindent \textbf{Language Understanding}. In the context of \ac{NLU}, \ac{LLMs} are employed to extract meaning from human language. \ac{LLMs} are being used for a variety of \ac{NLU}~\cite{mcshane2017natural} and other language-related tasks, including sentiment analysis and named entity recognition~\cite{brown2020language}. These models can analyze and comprehend the context of a given text, making them valuable for a wide range of applications.

\vspace{1.0ex}
\noindent \textbf{Machine Translation}. In the context of machine translation, \ac{LLMs} are used to automatically translate text between different languages~\cite{johnson2017google}. For example, Google Translate utilizes \ac{LLMs} to seamlessly translate text, documents, and websites from one language into another. This capability demonstrates the practical utility of \ac{LLMs} in bridging language barriers and enhancing communication, achieved through training on extensive multilingual datasets~\cite{brants2007large}. The quality of translation relies on the underlying capabilities of \ac{LLMs} for natural language understanding and generation. The work in~\cite{bahdanau2014neural} introduced the concept of attention to neural machine translation architecture, leading to significant advancements in language translation quality.

\vspace{1.0ex}
\noindent \textbf{Question Answering}. \ac{LLMs} are effectively employed in question-answering tasks across a variety of topics, enabling them to provide relevant answers to user queries~\cite{brown2020language, lewis2019bart}. This capability has applications in virtual assistants, information retrieval systems, and educational platforms. For example, the AI assistant from Google can answer questions about a variety of topics, such as current events, history, and science.

\vspace{1.0ex}
\noindent \textbf{Chatbots}. The \ac{NLP} capabilities of \ac{LLMs} contribute significantly to the development of intelligent chatbots~\cite{ouyang2022training}. This adaptability enhances the overall user experience, making interactions with virtual assistants more intuitive and effective. \ac{LLMs} are widely employed in creating chatbots for customer support and other interactive applications, enabling these intelligent virtual assistants to engage with humans, answer queries, and help in a natural and informative way~\cite{thoppilan2022lamda}. The ability of \ac{LLMs} to understand and respond to natural languages has opened up new possibilities in customer service, education, entertainment, and healthcare~\cite{thirunavukarasu2023large}. For example, companies like Facebook and Microsoft have successfully integrated \ac{LLMs} in their chatbot systems, such as Facebook's Messenger platform and Microsoft's Azure Bot Service. These platforms utilize the power of \ac{LLMs} to provide users with personalized and context-aware responses, demonstrating the practical applications of these models in real-world interactive environments.

\vspace{1.0ex}
\noindent \textbf{Speech Recognition}. Older speech recognition systems often relied on \ac{RNNs} or hybrid models combining \ac{HMMs} with \ac{DNNs}~\cite{hinton2012deep, graves2013speech}. However, these approaches faced limitations. \ac{RNNs} process input sequences one element at a time, leading to slow processing and difficulties handling long-range dependencies in audio signals~\cite{jozefowicz2016exploring}. Additionally, hybrid models were complex and required careful integration of separate components. To address these limitations, researchers have explored and applied \ac{LLMs} to speech recognition tasks, yielding promising results~\cite{shan2019investigating}. The core technology for speech recognition remains \ac{ASR} models specifically trained on vast amounts of speech data. These models excel at converting audio features into text but lack the broader language understanding capabilities of \ac{LLMs}. While traditional \ac{ASR} systems often rely on specialized architectures, the use of \ac{LLMs}, particularly transformer-based models, has gained attention for end-to-end speech recognition~\cite{vaswani2017attention}. \ac{LLMs} can analyze the output of \ac{ASR} models and suggest corrections based on their understanding of language and context, improving the accuracy of transcriptions, especially in noisy environments or with unclear pronunciations~\cite{salazar2019self}. Additionally, \ac{LLMs} can be leveraged to provide context to the speech recognition process. By considering surrounding text or information about the speaker and situation, \ac{LLMs} can assist \ac{ASR} models in making better decisions about what is being said.

\vspace{1.0ex}
\noindent \textbf{Text Summarization}. \ac{LLMs} have demonstrated successful applications in various text summarization tasks, such as summarizing documents and news articles~\cite{feng2018attentive, krantz2018abstractive}. For example, the work presented in~\cite{lewis2019bart} introduces a sequence-to-sequence pre-training model, which has proven highly effective in abstractive summarization tasks. Modern \ac{LLMs}, empowered with powerful \ac{NLP} capabilities, can understand the context of a document, and generate concise and coherent summaries quickly while preserving the overall meaning of the original text.

\vspace{1.0ex}
\noindent \textbf{Code Completion}. In addition to the capabilities of \ac{LLMs} to generate human-like text and perform various \ac{NLP} tasks, \ac{LLMs} have also demonstrated the ability to understand the context of code and generate relevant and accurate code suggestions~\cite{li2023starcoder}. Code completion with \ac{LLMs} involves predicting the next set of characters in a code snippet based on the provided context~\cite{svyatkovskiy2019pythia}. These models leverage their extensive pre-trained knowledge of programming languages and coding patterns to generate pertinent code suggestions~\cite{feng2020codebert}. This approach has been shown to improve developer productivity~\cite{lu2021codexglue}.

\section{Challenges of Generative AI and LLMs}
\label{challenges_of_generative_AI_and_LLMs}

Despite their wide range of immense potential for society, Generative \ac{AI} and \ac{LLMs} also pose several critical challenges that need to be carefully considered and addressed. These challenges include:

\vspace{1.0ex}
\noindent \textbf{Bias and Fairness}. One of the main challenges associated with Generative AI and \ac{LLMs} is the inheriting biases from the training data, which can lead to biased, unfair, and discriminatory outputs. Biased outputs from Generative \ac{AI} and \ac{LLMs} can have significant real-world consequences. For example, biased hiring algorithms may discriminate against certain job applicants. Potential bias problems like these can be mitigated by developing algorithms that are explicitly designed to be fair and unbiased by using approaches such as fairness-aware training~\cite{xu2018fairgan}, counterfactual analysis~\cite{feder2021causalm, chen2023disco, huang2019reducing}, and adversarial debiasing~\cite{zhang2018mitigating}.

\vspace{1.0ex}
\noindent \textbf{Interpretability}. Understanding and interpreting the decision-making process of \ac{LLMs} presents a significant challenge. The inherent lack of interpretability in these models raises serious concerns, especially in critical applications that require explainable decision-making. Addressing interpretability challenges in Generative \ac{AI} and \ac{LLMs} involves several approaches. One solution is to design \ac{LLMs} with inherent explainability features, such as employing interpretable model architectures and incorporating constraints that promote understandable decision-making. Another approach is to develop advanced techniques that provide insights into the inner workings of LLMs, such as saliency maps~\cite{ding2021evaluating}, attention mechanisms~\cite{vaswani2017attention}, and feature attribution methods. Additionally, implementing post-hoc interpretability methods~\cite{madsen2022post, kroeger2023large} including feature importance analysis and model-agnostic interpretation techniques, can offer valuable insights into the factors influencing the outputs of the model.

\vspace{1.0ex}
\noindent \textbf{Fine-tuning and Adaptability}. Fine-tuning large \ac{LLMs} for specific domains is challenging due to their inherent limitations in generalization. In addition to their limited ability to generalize, \ac{LLMs} may face difficulty in understanding and reasoning complex concepts, hindering their ability to adapt to new tasks. Addressing the challenges associated with fine-tuning and adaptability in Generative \ac{AI} and \ac{LLMs} involves exploring various approaches. One approach involves employing transfer learning techniques that leverage knowledge from pre-trained models on diverse datasets, allowing the model to capture a broader range of knowledge by accelerating learning and improving generalization~\cite{chronopoulou2019embarrassingly, you2020co}. Additionally, incorporating domain-specific data during fine-tuning can enhance the model's adaptability to particular tasks ensuring it learns domain-specific patterns and relationships. Incorporating symbolic reasoning capabilities into \ac{LLMs} can also enhance their ability to understand and manipulate abstract concepts~\cite{zhang2022elastic}. Leveraging meta-learning techniques to enable \ac{LLMs} to learn how to quickly learn also improves their ability to adapt to new tasks and data distributions~\cite{hou2022meta}.

\vspace{1.0ex}
\noindent \textbf{Domain Adaptation}. Most of the high-performing models being released are already fine-tuned for instruction-following. However, adapting these pre-trained \ac{LLMs}, which have been specifically fine-tuned for a specific domain (such as chat), to a new task (such as generating text formats or answering your questions) not formatted for instruction-following without compromising its performance in the original domain is challenging. The challenge lies in preserving the model's ability to understand and follow instructions while also enabling it to generate coherent and informative text in the new domain. This requires careful consideration of the training data, the model architecture, and the fine-tuning process.  However, fine-tuning \ac{LLMs} for an entirely new domain introduces the risk of negative transfer~\cite{wang2019characterizing}. This occurs when the model's new knowledge conflicts with its existing knowledge. Additionally, domain adaptation often requires access to a large amount of high-quality data from the new domain. This can be challenging to obtain, especially for specialized domains. Potential strategies for addressing this challenge include leveraging the weights of the pre-trained \ac{LLMs} as a starting point for the fine-tuning process, synthesizing additional data from the new domain to supplement the existing data, and simultaneous multi-task training involving both the original and new tasks.

\vspace{1.0ex}
\noindent \textbf{Data Privacy and Security}. \ac{LLMs} are trained on massive and diverse datasets that may contain sensitive personal information. The potential for unintentional disclosure of private or sensitive information during text generation is a significant concern. For instance, when applied in healthcare, the use of \ac{LLMs} raises concerns regarding patient privacy and the potential for misdiagnosis. There is also a risk of \ac{AI} systems being exploited for malicious purposes, such as generating fake identities, which raises privacy concerns. This, for example, has caused ChatGPT to be temporarily outlawed in Italy\footnote{\url{https://www.bbc.com/news/technology-65139406}}\footnote{\url{https://www.theverge.com/2023/4/28/23702883/chatgpt-italy-ban-lifted-gpdp-data-protection-age-verification}}. Addressing privacy concerns in Generative AI and \ac{LLMs} requires a multifaceted approach that includes enhancing model training with privacy-preserving techniques, such as federated learning, homomorphic encryption, or differential privacy~\cite{shokri2015privacy, abadi2016deep}.  Additionally, fine-tuning models on curated datasets that exclude sensitive information can help minimize the risk of unintentional disclosures. Ethical guidelines and regulations specific to AI applications, such as in healthcare, can provide further safeguards against privacy breaches~\cite{morley2021ethics, borenstein2021emerging}. \ac{LLMs} should be able to handle adversarial attacks, noisy data, and out-of-distribution inputs. In addition to this, it is worth mentioning that beyond model privacy, addressing concerns related to the privacy and security of the training and deployment data itself is important.

\vspace{1.0ex}
\noindent \textbf{Computational Cost}. Training and deploying \ac{LLMs} demand significant computational resources. This poses infrastructure challenges, energy consumption particularly for large-scale deployments, and accessibility of high-performance computing resources. As shown in Figure~\ref{Fig:time-line-LLMs}, the increase in model sizes comes with challenges related to computational requirements and resource accessibility. Reducing the computational cost of \ac{LLMs} involves several approaches. Firstly, optimizing model architectures and algorithms can enhance efficiency, reducing the computational burden without compromising performance. Secondly, leveraging distributed computing frameworks and specialized hardware accelerators, such as \ac{GPUs} or \ac{TPUs}, can significantly improve training speed and resource utilization~\cite{strubell2020energy}. In addition to this, employing quantization techniques~\cite{lin2023awq} to models that have already been trained\footnote{\url{https://huggingface.co/TheBloke}} is also important.

\vspace{1.0ex}
\noindent \textbf{Deepfake Generation}. Generative \ac{AI} models are widely used for deepfake generation~\cite{tolosana2020deepfakes}. Deepfakes utilize various generative models, including \ac{GANs}, to manipulate or generate realistic-looking content, primarily for image or video production~\cite{thies2016face2face, garrido2014automatic}. Despite their potential applications in various domains including education and entertainment, deepfakes also pose several potential risks due to their potential misuse, including the spread of misinformation and identity theft~\cite{brundage2018malicious}. The deepfakes technology can be exploited to create fake videos or audio recordings of individuals leading to the spread of misinformation or disinformation, which can have devastating consequences for individuals and society. It is, therefore, important to develop advanced techniques to mitigate the risks associated with deepfakes.

\vspace{1.0ex}
\noindent \textbf{Human-AI Collaboration}. \ac{LLMs} should be designed to enable seamless human-AI collaboration, enabling them to effectively understand and respond to human instructions and provide clear explanations for their outputs~\cite{adadi2018peeking}. To achieve effective human-AI collaboration, it is important to integrate humans into the design process of \ac{LLMs} to ensure that they are aligned with human needs and expectations~\cite{amershi2014power, mosqueira2023human}. To incorporate human feedback into the training process, we can utilize techniques such as \ac{RLHF}~\cite{griffith2013policy, macglashan2017interactive} and \ac{DPO}~\cite{rafailov2023direct} for training \ac{RL} agents using human feedback. Additionally, employing \ac{XAI} techniques for \ac{LLMs} can enhance the transparency and understandability of their decision-making processes~\cite{rosenfeld2019explainability}. Developing natural language interfaces that facilitate natural human-LLM interactions is another key aspect of enhancing human-AI collaboration~\cite{ribeiro2016should}. Conversational AI, intelligent chatbots, and voice assistants are examples of technologies that enable intuitive human-AI interactions.

\vspace{1.0ex}
\noindent \textbf{Long-Term Planning}. Generative models, particularly autoregressive models that generate text one token at a time, face challenges in long-term planning~\cite{valmeekam2023planning}. These models tend to focus on the immediate local context, making it difficult to maintain consistency over longer text passages. This limitation comes from the model's lack of a global view of the entire sequence it generates. Additionally, autoregressive models struggle to plan for situations with future uncertainties. To address the long-term planning challenge with \ac{LLMs}, we can employ several approaches including hierarchical attention, which allows \ac{LLMs} to focus on different parts of the input at different times that can help the models capture long-range dependencies~\cite{yang2016hierarchical}. Equipping \ac{LLMs} with memory that allows them to store information about the past, which can be used to inform future decisions, is another approach to address this challenge~\cite{grave2016improving}.

\vspace{1.0ex}
\noindent \textbf{Limited Context Window}. Having a limited context window is a fundamental challenge for \ac{LLMs} since they can only process a limited amount of text at a time. This limitation comes from their reliance on attention mechanisms~\cite{vaswani2017attention}, which allow them to focus on the most relevant parts of the text when generating content. The context window defines the number of tokens considered by the model during prediction, and a smaller context window can limit the model's ability to understand and generate a contextually relevant text, especially in long passages or documents. Several techniques can be employed to address the challenge of a limited context window. A common approach involves using hierarchical attention, which enables models to focus on different levels of context~\cite{yang2016hierarchical}. Additionally, the parallel context window approach allows for parallel processing of multiple context windows~\cite{ratner2023parallel}. This feature allows the models to store information beyond the immediate context window, enabling better handling of long-term dependencies~\cite{kaiser2017one}.

\vspace{1.0ex}
\noindent \textbf{Long-Term Memory}. \ac{LLMs} are trained on a massive corpus of text and code, but their completely stateless nature limits their ability to store and retrieve information from past experiences~\cite{ghodsi2020rnn}. This inherent lack of explicit memory restricts their ability to maintain context and engage in natural conversations, leading to less coherent responses, especially across multi-turn dialogues or tasks requiring information retention. Without the ability to remember past interactions, \ac{LLMs} cannot personalize their responses to specific users. This means they cannot adapt their communication style based on the user's preferences, interests, or previous interactions. Challenges associated with this limitation include issues of consistency and task continuity. To address these challenges, various approaches and techniques can be considered. Beyond context window techniques, integrating external memory mechanisms like memory networks or attention mechanisms with an external memory matrix can enhance the model's ability to access and update information across different turns~\cite{sukhbaatar2015end}. Alternatively, designing applications that externally maintain session-based context allows the model to reference past interactions within a session. Additionally, retrieval-based techniques enable \ac{LLMs} to access relevant information from past conversations or external sources during inference, enhancing their ability to maintain context and deliver more consistent responses~\cite{azerbayev2023llemma}.

\vspace{1.0ex}
\noindent \textbf{Measuring Capability and Quality}.  Traditional statistical quality measures, such as \emph{Accuracy} and \emph{F-score} do not easily translate to generative tasks~\cite{deutsch2022limitations}, especially long-form generative tasks. Furthermore, the accessibility of test sets in numerous benchmark datasets provides an avenue for the potential manipulation of leaderboards by unethical practitioners. This involves the inappropriate training of models on the test set, a practice likely employed by researchers seeking funding through achieving top positions on public leaderboards, such as Hugging Face's Open LLM Leaderboard\footnote{\url{https://huggingface.co/spaces/HuggingFaceH4/open_llm_leaderboard}}. At the time of writing this paper, a 7 billion parameter model is outperforming numerous 70 billion parameter models. A prospective and pragmatic approach to appraising model outputs is to utilize an auxiliary model for evaluating the generated content from the original model~\cite{zhu2023judgelm}. However, this methodology may prove ineffective if the judgment model lacks training within the specific domain it is employed to assess.

\vspace{1.0ex}
\noindent \textbf{A Concerning Trend Towards ``Closed'' Science}. As models transition from experimental endeavors to commercially viable products, there is a diminishing inclination to openly share the progress achieved within research laboratories\footnote{\url{https://www.theverge.com/2023/3/15/23640180/openai-gpt-4-launch-closed-research-ilya-sutskever-interview}}.  This shift poses a significant obstacle to the collaborative advancement of knowledge, hindering the ability to build upon established foundations when essential details are withheld. Furthermore, replicating published results becomes arduous when the prompts employed in the experimentation are not disclosed, since subtle alterations to prompts can, in some cases, significantly affect the performance of the model. Compounding these concerns, accessing the necessary resources to reproduce results often entails financial obligations to the publishers of the models, creating yet another barrier to entry into the scientific landscape for low-resource researchers. This situation prompts reflection on the current situation and the potential impediments it imposes on the pursuit of knowledge and innovation.

\section{Bridging Research Gaps and Future Directions}
\label{future_directions}
Our research has identified several key areas that require attention to ensure the ethical integration of Generative AI and LLMs. These areas include addressing issues such as bias and fairness in outputs, the necessity for models to provide explanations for their reasoning, and the challenges associated with adapting these models to diverse situations and domains. Furthermore, considerations regarding data privacy, security, and the potential for misuse in areas such as deepfakes require careful attention. Addressing these challenges through advancements in areas we have proposed, such as improved bias detection and the development of interpretable models, holds significant promise. Proactively tackling these issues is essential to ensuring that AI development is not only technically advanced but also beneficial to society. This includes developing clear metrics to assess model performance, enhancing their interpretability, and prioritizing user privacy and security. By incorporating ethical considerations into AI development, we pave the way for their responsible deployment across various domains, including healthcare, recruitment, and content creation. This will foster a future where AI serves as a positive force for societal good, promoting inclusivity and making a real impact.

\section{Conclusion}
\label{conclusion}

This paper explores the transformative potential of Generative \ac{AI} and \ac{LLMs}, highlighting their advancements, technical foundations, and practical applications across diverse domains. We argue that understanding the full potential and limitations of Generative \ac{AI} and \ac{LLMs} is crucial for shaping the responsible integration of these technologies. By addressing critical research gaps in areas such as bias, interpretability, deepfakes, and human-AI collaboration, our work paves the way for an impactful, ethical, and inclusive future of \ac{NLP}. We envision this research serving as a roadmap for the \ac{AI} community, empowering diverse domains with transformative tools and establishing a clear path for the responsible evolution of \ac{AI}. 

In our future work, we aim to explore advanced techniques for identifying and mitigating bias in both training data and algorithms to enhance fairness in AI systems. Additionally, we plan to investigate explainable AI approaches and develop new strategies to improve the interpretability of AI models. Building upon our previous line of research on human-autonomy teaming, we will delve into the development of models that facilitate seamless collaboration and interaction between humans and AI. We hope this work encourages researchers across multiple disciplines of the \ac{AI} community, from both academia and industry, to further explore the broader domain of Generative AI and LLMs.

\begin{comment}
\section*{Acknowledgment}

This work was supported by the DoD Center of Excellence in AI/ML at Howard University under Contract number W911NF-20-2-0277 with the US ARL and Mastercard Research Funds. However, any opinions, findings, conclusions, or recommendations expressed in this document are those of the authors and should not be interpreted as representing the official policies, either expressed or implied, of the funding agencies. 
\end{comment}

\bibliography{LLM_and_GAI}\addcontentsline{toc}{section}{\refname}
\bibliographystyle{IEEEtran} 
%\balance

%\begin{comment}

\begin{IEEEbiography}[{\includegraphics[width=1in,height=1.25in,clip,keepaspectratio]{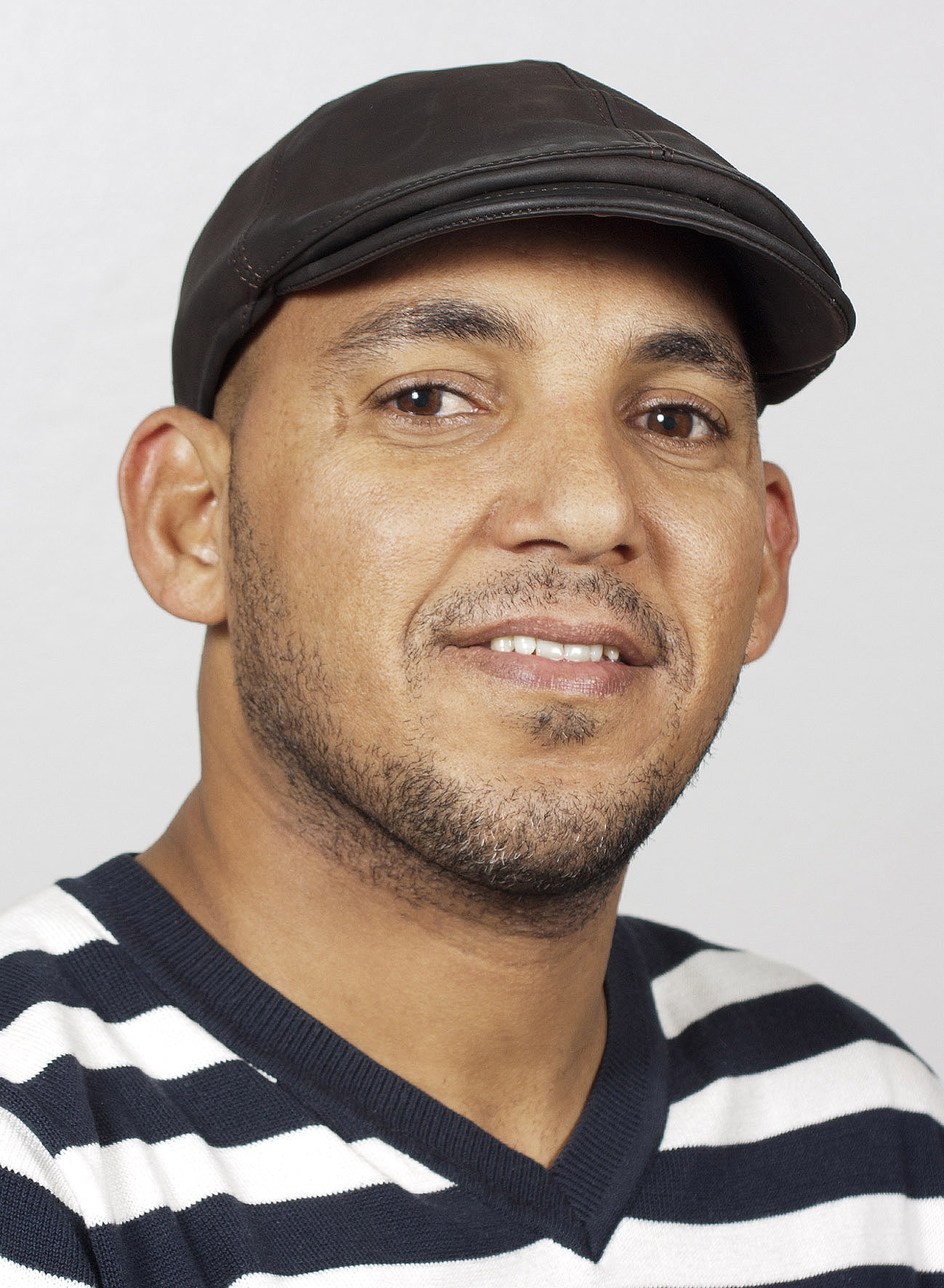}}]{Desta Haileselassie Hagos} (Member, IEEE) received a Ph.D. degree in Computer Science from the University of Oslo, Faculty of Mathematics and Natural Sciences, Norway, in April 2020. Currently, he is a Postdoctoral Research Fellow at the United States Department of Defense (DoD) Center of Excellence in Artificial Intelligence and Machine Learning (CoE-AIML), College of Engineering and Architecture (CEA), Department of Electrical Engineering and Computer Science at Howard University, Washington DC, USA. Previously, he was a Postdoctoral Research Fellow at the Division of Software and Computer Systems (SCS), Department of Computer Science, School of Electrical Engineering and Computer Science (EECS), KTH Royal Institute of Technology, Stockholm, Sweden, working on the H2020-EU project, ExtremeEarth: From Copernicus Big Data to Extreme Earth Analytics. He received his B.Sc. degree in Computer Science from Mekelle University, Department of Computer Science, Mekelle, Tigray, in 2008. He obtained his M.Sc. degree in Computer Science and Engineering specializing in Mobile Systems from Lule\aa ~University of Technology, Department of Computer Science Electrical and Space Engineering, Sweden, in June 2012. His current research interests are in the areas of Machine Learning, Deep Learning, and Artificial Intelligence.\end{IEEEbiography}

\begin{IEEEbiography}[{\includegraphics[width=1in,height=1.25in,clip,keepaspectratio]{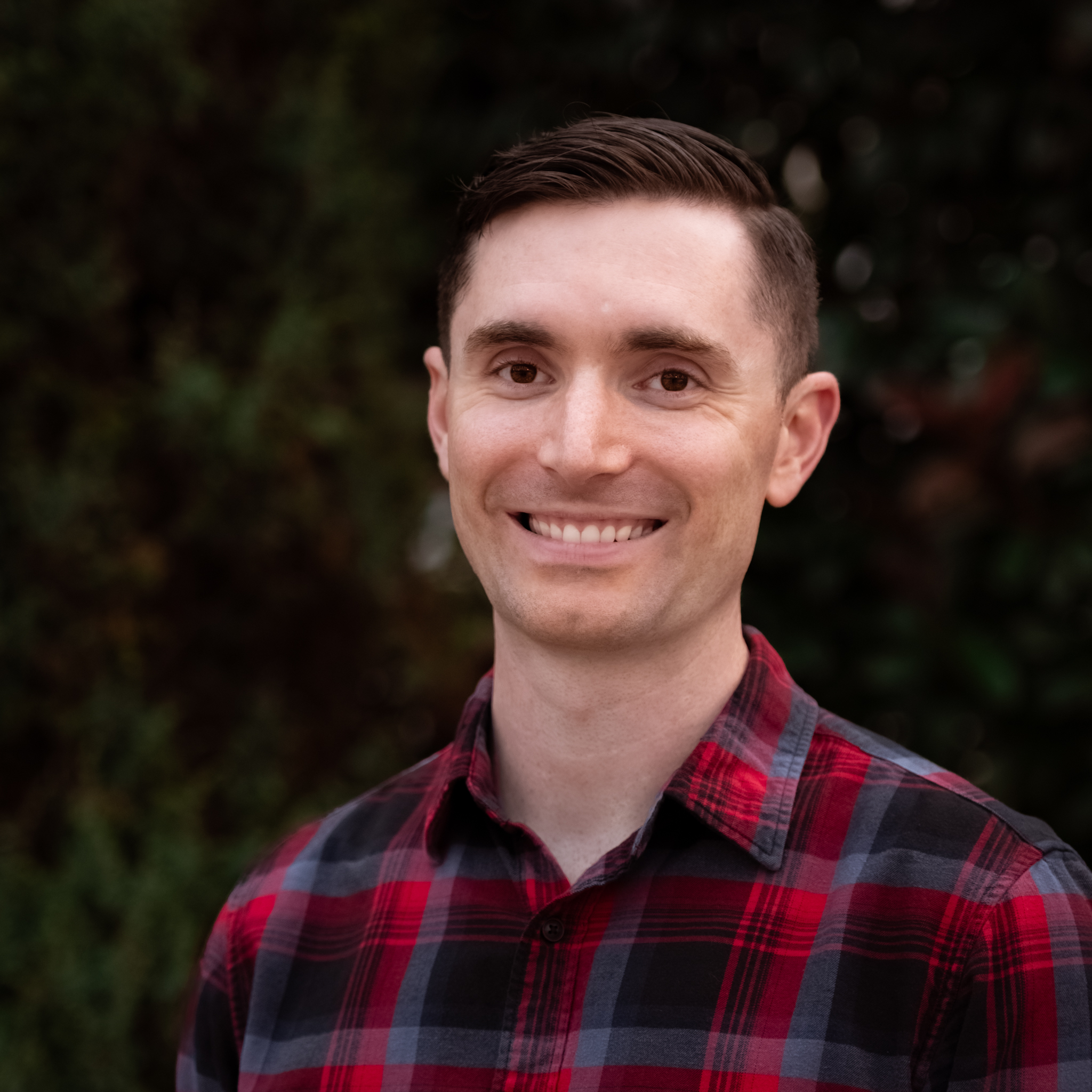}}]{Rick Battle} is a Staff Machine Learning Engineer at VMware by Broadcom.  He is the Head of NLP Research in VMware AI Labs.  He received a Master of Science degree in Computer Science with a specialization in Machine Learning from the Naval Postgraduate School in Monterey, CA, and earned a Bachelor of Science degree in Computer Engineering from Virginia Tech in Blacksburg, VA.  His research interests are in the areas of the application of Large Language Models to real-world use cases and Information Retrieval.
\end{IEEEbiography}

\begin{IEEEbiography}[{\includegraphics[width=1in,height=1.25in,clip,keepaspectratio]{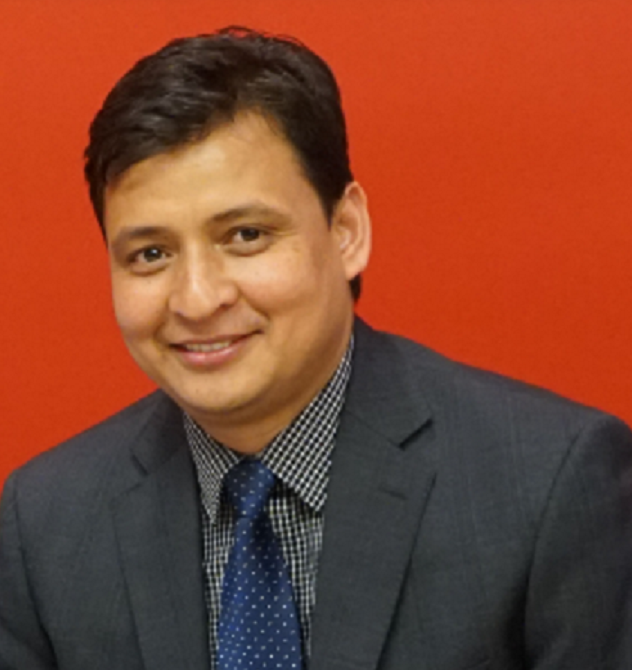}}]{Danda B. Rawat} (Senior Member, IEEE) is the Associate Dean for Research \& Graduate Studies, a Full Professor in the Department of Electrical Engineering \& Computer Science (EECS), Founding Director of the Howard University Data Science \& Cybersecurity Center, Founding Director of the DoD Center of Excellence in Artificial Intelligence \& Machine Learning (CoE-AIML), Director of Cyber-security and Wireless Networking Innovations (CWiNs) Research Lab at Howard University, Washington, DC, USA. Dr. Rawat is engaged in research and teaching in the areas of cybersecurity, machine learning, big data analytics, and wireless networking for emerging networked systems including cyber-physical systems (eHealth, energy, transportation), Internet-of-Things, multi-domain operations, smart cities, software-defined systems, and vehicular networks.Dr. Danda B. Rawat successfully led and established the Research Institute for Tactical Autonomy (RITA), the 15th  University Affiliated Research Center (UARC) of the US Department of Defense  as the PI/Founding Executive Director  at Howard University, Washington, DC, USA. Dr. Rawat is engaged in research and teaching in the areas of cybersecurity, machine learning, big data analytics and wireless networking for emerging networked systems including cyber-physical systems (eHealth, energy, transportation), Internet-of-Things, multi domain operations, smart cities, software defined systems and vehicular networks. Dr. Rawat has secured over \$110 million as a PI and over \$18 million as a Co-PI in research funding from the US National Science Foundation (NSF), US Department of Homeland Security (DHS), US National Security Agency (NSA), US Department of Energy, National Nuclear Security Administration (NNSA), National Institute of Health (NIH), US Department of Defense (DoD) and DoD Research Labs, Industry (Microsoft, Intel, VMware, PayPal, Mastercard, Meta, BAE, Raytheon etc.) and private Foundations. Dr. Rawat is the recipient of the US NSF CAREER Award, the US Department of Homeland Security (DHS) Scientific Leadership Award, Presidents’ Medal of Achievement Award (2023) at Howard University, Provost's Distinguished Service Award 2021,  the US Air Force Research Laboratory (AFRL) Summer Faculty Visiting Fellowship 2017, Outstanding Research Faculty Award (Award for Excellence in Scholarly Activity)and several Best Paper Awards. He has been serving as an Editor/Guest Editor for over 100 international journals including the Associate Editor of IEEE Transactions on Information Forensics \& Security,  Associate Editor of Transactions on Cognitive Communications and Networking, Associate Editor of IEEE Transactions of Service Computing, Editor of IEEE Internet of Things Journal, Editor of IEEE Communications Letters, Associate Editor of IEEE Transactions of Network Science and Engineering and Technical Editors of IEEE Network. He has been in Organizing Committees for several IEEE flagship conferences such as IEEE INFOCOM, IEEE CNS, IEEE ICC, IEEE GLOBECOM and so on. He served as a technical program committee (TPC) member for several international conferences including IEEE INFOCOM, IEEE GLOBECOM, IEEE CCNC, IEEE GreenCom, IEEE ICC, IEEE WCNC and IEEE VTC conferences. He served as a Vice Chair of the Executive Committee of the IEEE Savannah Section from 2013 to 2017. Dr. Rawat received the Ph.D. degree from Old Dominion University, Norfolk, Virginia in December 2010. Dr. Rawat is a Senior Member of IEEE and a Lifetime Professional Senior Member of ACM, a Lifetime Member of Association for the Advancement of Artificial Intelligence (AAAI), a lifetime member of SPIE, a member of ASEE and AAAS, and a Fellow of the Institution of Engineering and Technology (IET). He is an ACM Distinguished Speaker and an  IEEE Distinguished Lecturer (FNTC and VTS). 
\end{IEEEbiography}

%\end{comment}

\end{document}